\documentclass{article}

\usepackage{microtype}
\usepackage{graphicx}
\usepackage{subfigure}
\usepackage{paralist}
\usepackage{booktabs} 

\usepackage[ruled,vlined]{algorithm2e}

\usepackage{hyperref}
\usepackage{amsmath}
\usepackage{amssymb}
\usepackage{mathtools}
\usepackage{amsthm}
\usepackage{multicol}
\usepackage{multirow}

\usepackage[accepted]{icml2023}

\usepackage{amsmath}
\usepackage{amssymb}
\usepackage{mathtools}
\usepackage{amsthm}

\usepackage[capitalize,noabbrev]{cleveref}

\theoremstyle{plain}

\theoremstyle{definition}

\theoremstyle{remark}

\newcommand{\eg}[1]{\emph{e.g.}{#1}}

\usepackage[textsize=tiny]{todonotes}

\icmltitlerunning{Graph Switching Dynamical Systems}

\begin{document}

\twocolumn[
\icmltitle{Graph Switching Dynamical Systems}



\icmlsetsymbol{equal}{*}

\begin{icmlauthorlist}
\icmlauthor{Yongtuo Liu}{uva}
\icmlauthor{Sara Magliacane}{uva,mit-ibm}
\icmlauthor{Miltiadis Kofinas}{uva}
\icmlauthor{Efstratios Gavves}{uva}
\end{icmlauthorlist}

\icmlaffiliation{uva}{University of Amsterdam}
\icmlaffiliation{mit-ibm}{MIT-IBM Watson AI Lab}

\icmlcorrespondingauthor{Yongtuo Liu}{y.liu6@uva.nl}

\icmlkeywords{Machine Learning, ICML}

\vskip 0.3in
]



\printAffiliationsAndNotice{}  

\begin{abstract}
Dynamical systems with complex behaviours, e.g. immune system cells interacting with a pathogen, 
are commonly modelled by splitting the behaviour into different regimes, or \emph{modes}, each with simpler dynamics, 
and then learning the switching behaviour from one mode to another.
Switching Dynamical Systems (SDS) are a powerful tool that automatically discovers these modes and mode-switching behaviour from time series data.
While effective, these methods focus on \emph{independent objects}, where the modes of one object are independent of the modes of the other objects.
In this paper, we focus on the more general \emph{interacting object} setting for switching dynamical systems, where the per-object dynamics also depends 
on an unknown and dynamically changing subset of other objects and their modes.
To this end, we propose a novel graph-based approach for switching dynamical systems, GRAph Switching dynamical Systems (GRASS), in which we use a dynamic graph to characterize interactions between objects and learn both intra-object and inter-object mode-switching behaviour.
We introduce two new datasets for this setting, a synthesized ODE-driven particles dataset and a real-world Salsa Couple Dancing dataset. Experiments show that GRASS can consistently outperforms previous state-of-the-art methods.
\end{abstract}

\section{Introduction}
Complex time series are pervasive both in daily life and scientific research, usually consisting of sophisticated behaviours and interactions between entities or objects \citep{pavlovic2000learning, shi2021learning}. Consider for example emotion contagion in a crowd and how it might affect the crowd dynamics \citep{crowds}, or the differentiation of T cells, a crucial type of immune cell, into different subtypes with different roles after interacting with certain pathogens. 

A common way of modelling complex behaviour, e.g. represented by a discontinuous function, is by considering it as a sequence of simpler \emph{modes}, e.g. represented by a set of smooth functions. For example, the behaviour of a ball bouncing on the floor can be represented by two simple modes of falling and bouncing back. In many cases, the challenge is to identify the mode at each time point based on observations.
The state-of-the-art approaches for this task are Switching Linear Dynamical Systems (SLDS) \citep{ackerson1970state,ghahramani2000variational,oh2005variational} and their non-linear extensions, e.g. Switching Nonlinear Dynamical Systems (SNLDS) \citep{dong2020collapsed} and REDSDS \citep{ansari2021deep}. While effective, these approaches either model the mode of a single object, including modelling different objects as a ``super object'' \citep{dong2020collapsed,glaser2020recurrent}, or assume \emph{independent objects}, i.e. they model the mode of each object as independent from the others, e.g. dancing bees 
in \citep{ansari2021deep}.

\begin{figure*}[t]
 \centering
 \setlength{\tabcolsep}{1pt}
 \includegraphics[width=.90\linewidth]{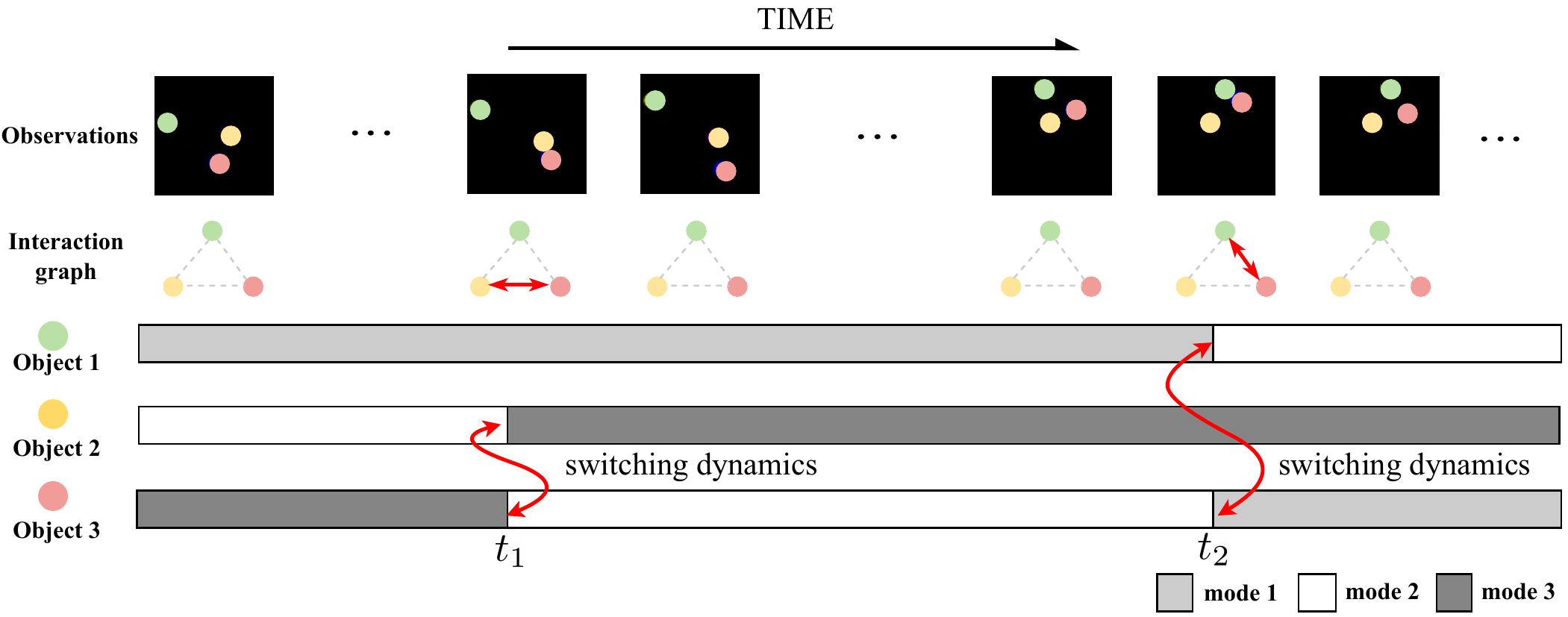}
  \vspace{-4pt}
 \caption{Illustration of Graph Switching Dynamical Systems (GRASS). As opposed to \emph{independent objects} Switching Dynamical Systems, where objects are processed independently, Graph Switching Dynamical Systems discover modes and mode-switching behaviours that can depend on object interactions. Interactions are modelled by a latent dynamic graph, which is inferred jointly with the other variables by maximizing the evidence lower bound.
 Activated interaction edges and mode switching are highlighted with red arrows, while inactive edges (no interactions) are visualized with grayed-out dashed lines in the interaction graph at each timestep.}
 \vspace{-2pt}
 \label{teaser}
 \end{figure*}

In this paper, we focus on the more general setting in which there are multiple \emph{interacting objects}, and in which the mode of an object can be influenced by the mode of the other objects. This is a more realistic setting for modelling many real-world systems, from crowds of people, to groups of immune cells and swarms. 
For this setting, we propose GRAph Switching dynamical Systems (GRASS) (described at high level in Fig.~\ref{teaser}), a framework that learns a dynamic graph to model interactions between objects and their modes across time, and can be combined with previously developed independent-objects switching dynamical systems methods.
To evaluate this new setting, we also propose two new datasets for benchmarking interacting object systems: a synthetic ODE-driven Particle dataset, and a Salsa Couple Dancing dataset, inspired by real-world benchmarks \citep{dong2020collapsed}.
Experiments show that GRASS outperforms the baselines and identifies mode-switching behaviors with higher accuracy and fewer switching errors.

\section{Multi Object Switching Dynamical Systems}
\label{sec:mosds}

We start from a collection of time series of observations $\mathbf{y} := \mathbf{y}_{1:T}^{1:N}$ for $T$ time steps and $N$ objects.
The $N$ objects move and their motions can be categorized to one out of $K$ possible \emph{ modes}.
For instance, an object might be moving in a spiral trajectory (mode 1) or it might bouncing on a wall (mode 2).
The $N$ objects interact with each other, and their motions change according to these interactions.
For instance, after a collision, an object might switch from a spiral to a sinusoidal motion.
The dynamics of these objects are governed by three types of variables: \emph{mode variables}, \emph{count variables} and \emph{state variables}. Mode variables are categorical variables $\mathbf{z} := \mathbf{z}_{1:T}^{1:N}=\{z_t^1, \dots, z_t^N\}_{t=1}^T$,  where $ z_t^n \in \{ 0, \dots, K-1\}$ denotes the mode for each time step $ t \in (1, \dots, T)$ and for each object $ n \in (1, \dots, N)$.
For instance, $z_{t=10}^{n=2}=3$ and $z_{t=10}^{n=5}=4$ mean that, at time step 10, the second object moves according to the third dynamic mode (for instance a spiral trajectory), while the fifth object moves according to the fourth dynamic mode (for instance a sinusoidal trajectory). Count variables are categorical variables $\mathbf{c} := \mathbf{c}_{1:T}^{1:N} = \{c_t^1, \dots, c_t^N\}_{t=1}^T$, where each $c_t^n \in (1, \dots, M)$ explicitly models the durations between switching modes for each object $n$ and each timestep $t$ and $M$ is the maximum number of steps before a switch.
These variables help us avoid frequent mode switching, caused by the fact that durations typically follow a geometric distribution, biasing unfavourably towards shorter durations ~\citep{ansari2021deep}. State variables are continuous variables $\mathbf{x} := \mathbf{x}_{1:T}^{1:N} = \{\mathbf{x}_t^1, \dots, \mathbf{x}_t^N\}_{t=1}^T$, where each $\mathbf{x}_t^n \in \mathbb{R}^d$ encodes the dynamics content per object and time step.
For instance, at time step $t$, $\mathbf{x}_t^n$ could encode the position and velocity of the trajectory of the $n$-th object.

\subsection{Interactions between all objects}

We formulate a probabilistic graphical model to describe our system of multiple interacting objects.
We first start with a formulation in which the modes of each object are affected by the modes of all other objects.
Then, in Section~\ref{sec:inference} we extend our system within a dynamic graph, with which we can learn at which time steps there exist interactions and between which objects as described in Section~\ref{sec:grass}.
Assuming Markovian dynamics and extending the standard Switching Dynamical Systems~\citep{linderman2016recurrent,ansari2021deep} paradigm to the case of $N$ objects, we assume the joint probability distribution is
\begin{align}
p(\mathbf{y}, \mathbf{x}, \mathbf{z}, \mathbf{c}) = 
\underbrace{\prod_{n=1}^N p(\mathbf{y}_1^n|\mathbf{x}_1^{n})
\,p(\mathbf{x}_1^n|z_1^{n})\,p(z_1^n)}_{\rm{Initial\,\,States}} \cdot \notag \\
\underbrace{\prod_{t=2}^{T} p(\mathbf{z}_{t}^{1:N}|\mathbf{z}_{t-1}^{1:N}, \mathbf{x}_{t-1}^{1:N}, \mathbf{c}_{t}^{1:N})}_{\rm{Interacting\,\, Modes}} \cdot  \notag \\ 
 \prod_{n=1}^{N} 
 \underbrace{
 \prod_{t=2}^{T}\left( 
 p(\mathbf{y}_t^{n}|\mathbf{x}_t^{n})
p(\mathbf{x}_t^{n}|\mathbf{x}_{t-1}^{n}, z_{t}^{n})
p(c_{t}^{n}|z_{t-1}^{n}, c_{t-1}^{n})\!\right)
}_{\rm{Per-object \,\, dynamics}}
\label{eq:joint-full}
\end{align}
We start by describing the per-object dynamics. In this case, we model for each object $n$ an \emph{observation probability} $p(\mathbf{y}_t^{n}|\mathbf{x}_t^{n})$, a state transition probability $p(\mathbf{x}_t^{n}|\mathbf{x}_{t-1}^{n}, z_{t}^{n})$ and a count transition probability $p(c_{t}^{n}|z_{t-1}^{n}, c_{t-1}^{n})$.
The observation probability $p(\mathbf{y}_t^{n}|\mathbf{x}_t^{n})$  models how the continuous state variables for this object $\mathbf{x}_t^n$ map into the observations $\mathbf{y}_t^n$. The state transition probability $p(\mathbf{x}_t^{n}|\mathbf{x}_{t-1}^{n}, z_{t}^{n})$ models how the continuous state variables at time $t$ are influenced by their previous values at time $t-1$ conditioned on mode variable for this object $z_{t}^{n}$. The count transition probability $p(c_{t}^{n}|z_{t-1}^{n}, c_{t-1}^{n})$ models how the count variables at time $t$ depend on their previous values at time $t-1$ and on the mode for this object at the previous time step $z_{t-1}^{n}$. The initial states have a similar setup, but in this case the state transition probability does not have an input from the previous timestep and the count variables are initialized at 1.
The mode transition probability $p(\mathbf{z}_{t}^{1:N}|\mathbf{z}_{t-1}^{1:N}, \mathbf{x}_{t-1}^{1:N}, \mathbf{c}_{t}^{1:N})$ models how the modes of objects are affected by the modes of all other objects $\mathbf{z}_{t-1}^{1:N}$, conditioned on the state variables $\mathbf{x}_{t-1}^{1:N}$ and count variables $\mathbf{c}_{t}^{1:N}$. 
In the absence of any knowledge on what interactions take place, this probability considers that all objects may potentially influence all other objects.

%
In Eq.~\eqref{eq:joint-full} except for the mode transition probability in the Interacting Modes term, all other terms $p(\mathbf{y}^n_1|\mathbf{x}^n_1)$, $p(\mathbf{x}^n_1|\mathbf{z}^n_1)$, $p(\mathbf{z}^n_1)$, $p(\mathbf{y}^n_t|\mathbf{x}^n_t)$, $p(\mathbf{x}^n_t|\mathbf{x}^n_{t-1}$, $\mathbf{z}^n_{t})$, $p(\mathbf{c}^n_{t}|\mathbf{z}^n_{t-1}, \mathbf{c}^n_{t-1})$ 
are factorized per object and thus similar independent-object dynamical systems treating all $N$ objects independently.
We refer to~\citep{dong2020collapsed, ansari2021deep} for details.

\subsection{Learning an amortized transition dynamics}

To simplify the modelling of switching dynamics, we assume that current dynamics for each object at time $t$ is independent from other objects given the complete latent state at $t-1$.
By further adopting a mixture representation for the marginal transition probabilities~\citep{raftery1985model, saul1999mixed}, we assume we can explicitly model pairwise mode-to-mode and object-to-object effects:
\begin{align}
&p(\mathbf{z}_{t}^{1:N}|\mathbf{z}_{t-1}^{1:N}, \mathbf{x}_{t-1}^{1:N},\mathbf{c}_{t}^{1:N})
\!=\prod_{n=1}^{N}p(z_t^n|\mathbf{z}_{t-1}^{1:N}, \mathbf{x}_{t-1}^{1:N}, c_{t}^{n}) \nonumber
 \!\\&=
\prod_{n=1}^{N}\sum_{m=1}^{N} w_t^{m \rightarrow n}\,p(z_t^n|z_{t-1}^{m}, \mathbf{x}_{t-1}^{m,n}, c_{t}^{n}),
\label{eq:transition-factorized-weighed}
\end{align}
where $\mathbf{x}_{t-1}^{m,n}=f_e(\mathbf{x}_{t-1}^m, \mathbf{x}_{t-1}^n)$ is a representation that aggregates the continuous  states of objects $m$ and $n$, for instance concatenation and $w_t^{m \rightarrow n}$ is the \emph{local dynamic factor} for objects $m$ and $n$, which satisfies $w_t^{m \rightarrow n}\geq 0$ and $\sum_{m=1}^{N} w_t^{m \rightarrow n}=1$.
This mixture assumption implies that the dynamics of object $m$ at time $t$ depends only on pairwise interactions with all other objects $n=1, \dots, N$, at time $t-1$, ignoring higher-order interactions.
The local dynamic factors allow dropping interactions between objects when none exist, since in multi-object systems, objects often affect one another at sparse points in time and space. 

The amortized transition dynamics benefits our modelling, because they allow us to model a larger number of objects and their switching dynamics (whether there exist or not) by simply extending the respective products and sums.
In the next section, we show how we can learn and use these local dynamic factors to ensure interaction sparsity more effectively when we learn a dynamic graph.


\section{Graph Switching Dynamical Systems}
\label{sec:grass}

Since our system consists of multiple objects, which may or may not interact at random points in time, we can model the system with a dynamic graph $\mathcal{G}_t=(\mathcal{V}_t, \mathcal{E}_t)$, whose structure and information varies across time.
The nodes $\mathcal{V}_t$ are all latent variables and observations related to each object $m$ at time step $t$, that is $\mathbf{v}^m_t=\{\mathbf{z}^m_t, \mathbf{x}^m_t, \mathbf{y}^m_t, \mathbf{c}^m_t\} \in \mathcal{V}_t$.
The edges $\mathbf{e}_t^{m\rightarrow n} \in \mathcal{E}_t$ denote whether there is an interaction between objects $m$ and $n$ at time $t$, which include self loops, i.e., $\mathbf{e}_t^{m\rightarrow m} \in \mathcal{E}_t , \forall m \in (1, \dots, N)$.

\begin{figure*}[t]
 \centering
 \setlength{\tabcolsep}{1pt}
 \includegraphics[width=.80\linewidth]{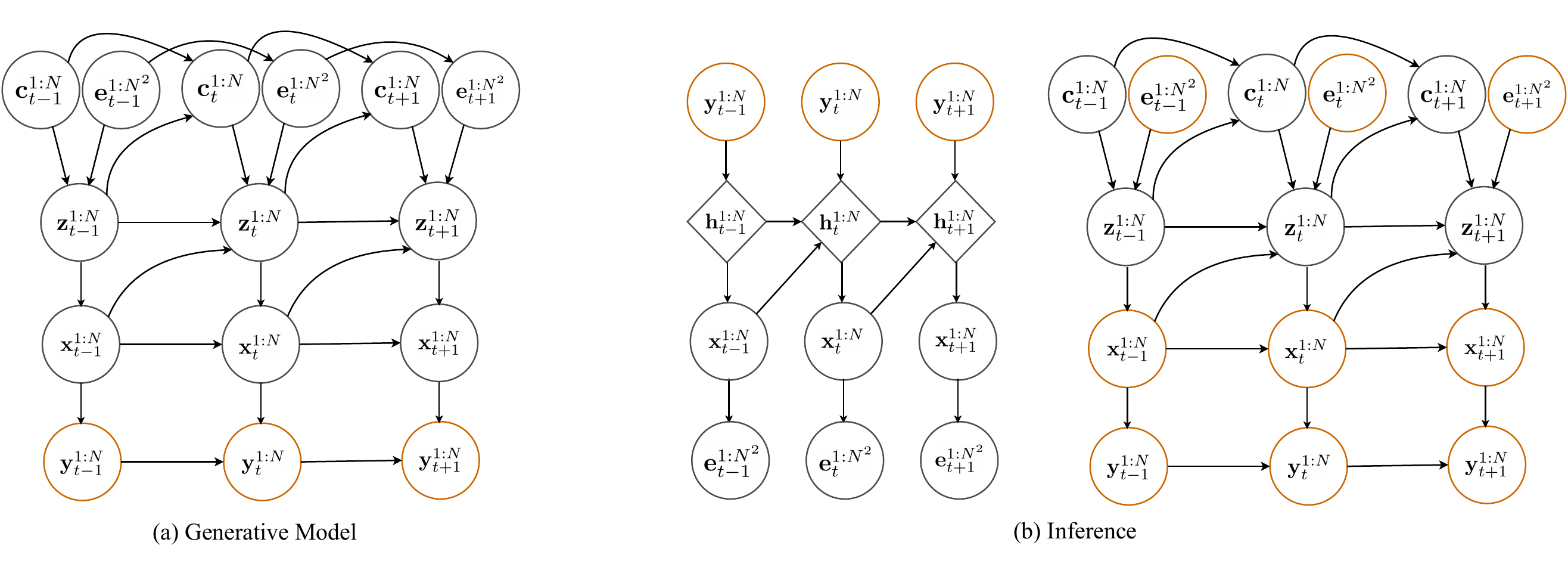}
 \vspace{-10pt}
 \caption{(a) Generative model of GRASS. (b) Left: Amortized approximate inference for the continuous states $\mathbf{x}_{t}^{1:N}$ and discrete edge variable $\mathbf{e}_{t}^{1:N^2}$ by inference networks. Temporal dependence is modeled by an intermediate latent embedding $\mathbf{h}_{t}^{1:N}$ which is given by directional RNNs. Right: Exact inference of discrete mode and count variables $\mathbf{z}_{t}^{1:N}$ and $\mathbf{c}_{t}^{1:N}$ based on the approximate pseudo-observations and pseudo-interactions $\mathbf{x}_{t}^{1:N}$ and $\mathbf{e}_{t}^{1:N^2}$. Orange circles denote observations or approximate pseudo-observations.}
 \label{whole model}
 \end{figure*}

Embedding the switching dynamical system into a graph topology, we want messages to be passed between graph nodes $\mathbf{v}^m$ and $\mathbf{v}^n$ via edges to signal interactions between objects.
Since we cannot know when interactions take place, how do they take place, and between what objects, we set the latent edge variables to be one-hot vectors of $L+1$ dimensions, $\mathbf{e}^{m\rightarrow n}_t=[e^{m\rightarrow n}_{t, 1}, ..., e^{m\rightarrow n}_{t, L+1}]$, where $e^{m\rightarrow n}_{t, l} \in \{0, 1\}$.
Setting the $l$-th dimension to 1, $e^{m\rightarrow n}_{t, l}=1$, indicates the $l$-th type of interaction is active between objects $m$ and $n$ at time $t$, with $e^{m\rightarrow n}_{t, l=1}=1$ standing for ``no interaction''.
Further, we set the prior edge distribution $p_{\theta}(\mathbf{e}_t) = \prod_{m\neq n}p_{\theta}(\mathbf{e}_t^{m\rightarrow n})$ to be a factorized object-to-object uniform distribution over edge types. We set the prior probability to be higher for ``no interaction'' edges, thus encouraging sparse graphs.

We enable two types of messages to be passed via the edges.
First, we want the latent edges to signal whether there is an interaction between two objects.
Thus, for objects $m$ and $n$ we set the unnormalized local dynamic factor $\tilde{w}_t^{m \rightarrow n}$ to be the sum of $L$ possible types of interaction:
\begin{equation}
\tilde{w}_t^{m \rightarrow n} = \sum_{l=2}^{L+1} e^{m\rightarrow n}_{t, l},
\qquad 
w_t^{m \rightarrow n} = \frac{\tilde{w}_t^{m \rightarrow n}}{\sum_{m=1}^N \tilde{w}_t^{m \rightarrow n}}
\end{equation}
Note that since the count starts from $l=2$ ($l=1$ stands for no interaction), $\tilde{w}_t^{m \rightarrow n}$ sums up to either 0 (no interaction) or 1.
$\tilde{w}_t^{m \rightarrow n}$ is a local influence weight
from object $m$ to object $n$. For the local dynamic factor, we normalize the weights over $m$ to get the weighted influence from all $m$ to $n$ that we use in the interacting modes term of Eq.~\eqref{whole model}.

We also want the edges to influence how the continuous state of a pair of objects $\mathbf{x}_{t-1}^{m,n}$ changes in case of an interaction.
To attain this, rather than simply concatenating features in $\mathbf{x}_{t-1}^m$ and $\mathbf{x}_{t-1}^n$ in Eq.~\eqref{eq:transition-factorized-weighed}, we use the edges as weights:
\begin{align}
\mathbf{x}^{m, n}_{t-1} = \sum_{l} e^{m\rightarrow n}_{t, l}\cdot f_e^l([\mathbf{x}_{t-1}^{m}, \mathbf{x}_{t-1}^{n}]),
\label{eq:interaction-induced continuous states equation}
\end{align}
where $f_e^l$ means a function for edge type $l$ that aggregates continuous states between any object pair into a single representation. These $L$ functions represent different interaction types indexed by the edge type $l=2, \dots, L+1$, similar to \citet{kipf2018neural}. Note that there is no need for a specific function for the `no interaction' case.

Taking into account the latent edge variables that are part of our probabilistic model, the joint probability becomes:
\begin{align}
p(\mathbf{y}, \mathbf{x}, \mathbf{z}, \mathbf{c}, \mathbf{e}) = 
\underbrace{\prod_{n=1}^N p(\mathbf{y}_1^n|\mathbf{x}_1^{n})
\,p(\mathbf{x}_1^n|z_1^{n})\,p(z_1^n)}_{\rm{Initial\,\,States}} \cdot \notag \\
\underbrace{\prod_{t=2}^{T} \prod_{n=1}^{N}\!\sum_{m=1}^{N}\!\!w_t^{m\rightarrow n}\,p(z_{t}^{n}|z_{t-1}^{m}, \mathbf{x}_{t-1}^{m,n}, c_{t}^{n}, \mathbf{e}_t^{m\rightarrow n})}_{\rm{Pairwise \, \, Interacting\,\, Modes}} \cdot  \notag \\ 
 \prod_{n=1}^{N} 
 \underbrace{
 \prod_{t=2}^{T}\left( 
 p(\mathbf{y}_t^{n}|\mathbf{x}_t^{n})
p(\mathbf{x}_t^{n}|\mathbf{x}_{t-1}^{n}, z_{t}^{n})
p(c_{t}^{n}|z_{t-1}^{n}, c_{t-1}^{n})\!\right)
}_{\rm{Per-object \,\, dynamics}},
\label{eq:grass}
\end{align}
The overall generative model and inference stages of GRASS are detailed in Fig.~\ref{whole model}. We show a more detailed version with the complete factorization in App.~\ref{appendix: Implementation Details}.

\section{Neural Network Implementation}
\label{sec:graph}

We use neural networks to model the terms in the joint likelihoods of our Switching Dynamical Systems, specifically of Eq.~\eqref{eq:joint-full} for the Multiple-Object Switching Dynamical System (MOSDS) of Section~\ref{sec:mosds}, and
of Eq.~\eqref{eq:grass} for Graph Switching Dynamical Systems (GRASS) of  Section~\ref{sec:grass}.

Since the mode variables $\mathbf{z}_t^{1:N}$ take one out of $K$ possible values for dynamic modes, we model them as categorical variables, parameterized by transition probabilities $T_t$.
Specifically, for pairs of objects in our system, we have:
\begin{align}
\!\! \!  p(z_{t}^{n}|z_{t-1}^{n}, \mathbf{x}_{t-1}^{m,n}\!, c_{t}^{n}, \mathbf{e}_t^{m\rightarrow n}\!)\!=\!\!
\begin{cases}\!
\delta_{z_{t}^{n}=z_{t-1}^{n}}    &  \!\!\!\! {\rm if}\;c_{t}^{n} > 1 \\
{\rm Cat}(z_{t}^{n};  T_t )  & \!\!\!\!{\rm if}\;c_{t}^{n} = 1
\! \end{cases}
\label{eq:transition-categorical}
\end{align}
where we resample the dynamic modes of objects or preserve them via a Kronecker $\delta$ function depending on whether our count variable is reset or not.

For MOSDS, we model the parameters $T_t$ of the categorical distributions in Eq.~\eqref{eq:transition-categorical} with a neural network $T_t=f_{z}(\mathbf{x}_{t}^{1:N})$ that takes as input the continuous states of all objects.
In this case, the neural network returns a $NK \times NK$ transition matrix per time step $t$, where rows correspond to past modes $\mathbf{z}^{1:N}_{t-1}$ and columns correspond to current modes $\mathbf{z}^{1:N}_t$.
The shape of the matrix $NK \times NK$ is because the neural network must predict in one forward pass the likelihoods for all possible combinations of (object $m$, object $n$, mode $i$, mode $j$).
Clearly, such a neural network is prohibitively expensive as it scales exponentially with the number of objects $N$ and modes $K$, and also wasteful to optimize, as it assumes object pairs do not share any dynamics at all.
So for GRASS, we instead model the parameters $T_t$ in Eq.~\eqref{eq:transition-categorical} with an amortized neural network $T_t=f_z^{l}(\mathbf{x}_{t-1}^{m,n})$ that takes as input only pairs of continuous states (the weights of the neural network are shared for any pair of objects).

For both MOSDS and GRASS, the neural network $f_z$ is a simple MLP. 
To satisfy the positivity $T_{t, i, j} > 0 \; \forall i, j=1,..., K$ and $\ell_1$ constraints $\sum_{j} T_{t, i, j}=1 \; \forall i=1,..., K$ for $T_t$, we apply a tempered softmax on $f_z$, $\mathcal{S}_{\tau} \circ f_z(\cdot)$.
The latent edges also take one out of $L+1$ possible values for different types of interactions.
Thus, we model them by an $L+1$-way categorical distribution as well.

\subsection{Inference}
\label{sec:inference}

Due to the exponential complexity of the state space, exact inference of latent variables in Switching Dynamical Systems is intractable.
Similar to~\citet{ansari2021deep}, we resort to approximate variational inference with neural networks for the continuous latent variable.
Furthermore, we modify the original forward-and-backward algorithm by \citet{yu2010hidden} to perform exact inference for the discrete mode and count variables, as we will detail below.
The variational approximation of the true posterior 
is $p(\mathbf{x},\mathbf{e},\mathbf{z},\mathbf{c}\,|\,\mathbf{y}) \approx q(\mathbf{x},\mathbf{e}\!,\mathbf{z},\!\mathbf{c}\,|\,\mathbf{y}) = q_{\phi_x}\!(\mathbf{x}|\mathbf{y})  q_{\phi_e}\!(\mathbf{e}|\mathbf{x})\,p_{\theta}(\mathbf{z},\mathbf{c}|\mathbf{y},\mathbf{x},\mathbf{e})$.
The $q_{\phi_x}$ and $q_{\phi_e}$ correspond to neural networks for the approximate inference of the continuous state and discrete edge variables, respectively, and parameterized accordingly.
We now describe the exact and approximate inference for each variable.
To summarize our setup, we provide a flowchat of the inference algorithm of GRASS in App.~\ref{appendix: Inference Algorithm}. The network architecture and implementation details are in App.~\ref{appendix: Implementation Details}.

\paragraph{Approximate inference of continuous state $\mathbf{x}$.} 
Following~\citep{dong2020collapsed,ansari2021deep}, we factorize the approximate posterior of $\mathbf{x}$ as $q_{\phi_x}\!(\mathbf{x}_{1:T}^{1:N}|\mathbf{y}_{1:T}^{1:N})=\prod_{n=1}^{N} q_{\phi_x}\!(\mathbf{x}_{1:T}^{n}|\mathbf{y}_{1:T}^{n})$.
In particular, we first process observations $\mathbf{y}_{1:T}^{n}$ by a bi-RNN to accumulate temporally smoothed embedding $\mathbf{h}_{1:T}^{n}$.
Then, we feed the embedding of the bi-RNN into a causal (\emph{i.e.} forward uni-directional) RNN, which outputs the overall posterior distribution $q_{\phi_x}\!(\mathbf{x}_{1:T}^{1:N}|\mathbf{y}_{1:T}^{1:N}) = \prod_{n}\prod_{t}q_{\phi_x}\!(\mathbf{x}_t^n|\mathbf{x}_{1:t-1}^n, \mathbf{h}_{1:t-1}^n)$.

\paragraph{Approximate inference of discrete edge $\mathbf{e}$.}
Given the inferred $\tilde{\mathbf{x}}_{1:T}^{1:N}\sim q_{\phi_x}\!(\mathbf{x}_{1:T}^{1:N}|\mathbf{y}_{1:T}^{1:N})$, we next infer the latent interaction graph structure of our graph $\mathcal{G}_t$.
We use a graph neural network $f_{\phi_z}(\tilde{\mathbf{x}}_{1:T}^{1:N})$, which is potentially fully connected and with self loops, where the node embeddings are the sampled continuous states $\tilde{\mathbf{x}}_t^m$.
We obtain relational edge embeddings $\mathbf{h}_{m\rightarrow n}^2$ by two rounds of message passing:
\begin{align}
                      &\mathbf{h}_{m}^1 = f_{\phi_z}^{\rm emb}(\tilde{\mathbf{x}}_t^{m}) \\
v\rightarrow e\!: \,\,&\mathbf{h}_{m\rightarrow n}^1 = f_{\phi_z}^{e,1}([\mathbf{h}_{m}^1, \mathbf{h}_{n}^1]) \\
e\rightarrow v\!: \,\,&\mathbf{h}_{m}^2 = f_{\phi_z}^{v,1}(\sum_{n=1}^N \mathbf{h}_{n\rightarrow m}^1) \\
v\rightarrow e\!: \,\,&\mathbf{h}_{m\rightarrow n}^2 = f_{\phi_z}^{e,2}([\mathbf{h}_{m}^2, \mathbf{h}_{n}^2])
\end{align}

Assuming conditional independence between edges given all the inferred states, the approximate posterior for edge types becomes $q_{\phi_e}\!(\mathbf{e}_{1:T}^{1:N^2}|\tilde{\mathbf{x}}_{1:T}^{1:N}) = \prod_{t}q_{\phi_e}\!(\mathbf{e}_{t}^{1:N^2}|\tilde{\mathbf{x}}_{1:t}^{1:N})=\prod_{t}\prod_{m,n}q_{\phi_e}\!(\mathbf{e}_{t}^{m\rightarrow n}|\tilde{\mathbf{x}}_{1:t}^{1:N})=\prod_{t}\prod_{m,n}{\rm softmax}((\mathbf{h}_{m\rightarrow n}^2 + \mathbf{g})/\tau)$, where $\mathbf{g}$ is a vector sampled from a ${\rm Gumbel}(0,1)$ distribution for the reparametrization trick and $\tau$ is a temperature to control relaxation smoothness~\citep{maddison2016concrete}.

\paragraph{Exact inference of discrete mode $\mathbf{z}$ and count $c$.}
Given the inferred states $\tilde{\mathbf{x}}_{1:T}^{1:N}\sim q_{\phi_x}\!(\mathbf{x}_{1:T}^{1:N}|\mathbf{y}_{1:T}^{1:N})$ and edges $\tilde{\mathbf{e}}_{1:T}^{1:N^2}\sim q_{\phi_e}\!(\mathbf{e}_{1:T}^{1:N^2}|\tilde{\mathbf{x}}_{1:T}^{1:N})$, we do exact inference of the discrete mode and count variables $p_{\theta}(\mathbf{z}_{1:T}^{1:N},\mathbf{c}_{1:T}^{1:N}|\mathbf{y}_{1:T}^{1:N},\tilde{\mathbf{x}}_{1:T}^{1:N},\tilde{\mathbf{e}}_{1:T}^{1:N^2})$. We modify the forward-backward algorithm used with hidden Markov models~\citep{collins2013forward} by introducing the additional continuous state $\tilde{\mathbf{x}}_{1:T}^{1:N}$ and discrete edge $\tilde{\mathbf{e}}_{1:T}^{1:N^2}$ variables, where the forward part $\alpha_t$ and backward part $\beta_t$ are defined as:
\begin{align}
\alpha_t(\mathbf{z}_{t},\mathbf{c}_{t}) &= p(\mathbf{y}_{1:t},\tilde{\mathbf{x}}_{1:t},\tilde{\mathbf{e}}_{1:t}, \mathbf{z}_{1:t},\mathbf{c}_{1:t}) \\
\beta_t(\mathbf{z}_{t},\mathbf{c}_{t})
&= p(\mathbf{y}_{t+1:T},\tilde{\mathbf{x}}_{t+1:T}\,|\,\tilde{\mathbf{x}}_{t},\tilde{\mathbf{e}}_{t}, \mathbf{z}_{t},\mathbf{c}_{t}),
\end{align}
where we drop for clarity superscripts from $\mathbf{z}_{t}, \mathbf{c}_{t}, \mathbf{y}_{t}, \mathbf{x}_{t}$, and $\tilde{\mathbf{e}}_{t}$.
We describe the details in App.~\ref{app:fwbw}.

\begin{figure*}[t]
 \centering
 \setlength{\tabcolsep}{1pt}
 \includegraphics[width=.95\linewidth]{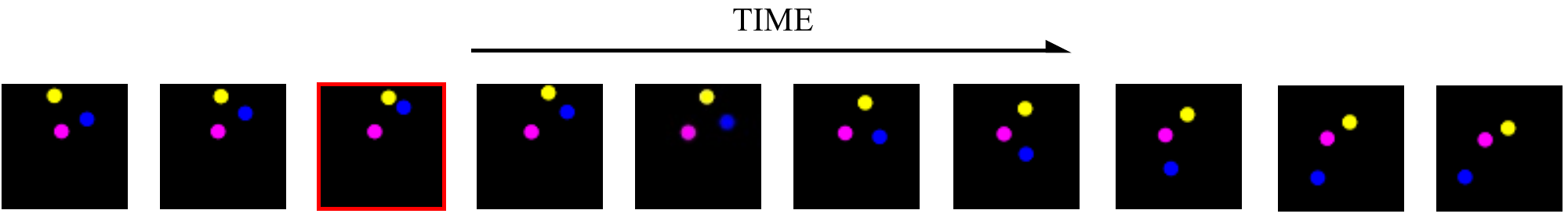}
  \vspace{-8pt}
 \caption{Visualization of ODE-driven particle dataset. Yellow and blue ball in the third frame switch their equations when they collide.}
 \vspace{-8pt}
 \label{ODE-figure}
 \end{figure*}

\subsection{Learning}
The overall network is jointly learned by maximizing the evidence lower bound~\citep{kingma2013auto},
\begin{align}
&\,\,{\rm log}\,p_\theta(\mathbf{y})\!-\!D_{K\!L}\left[q_\phi(\mathbf{x},\mathbf{z},\mathbf{c},\mathbf{e}|\mathbf{y})\,\|\,p_\theta(\mathbf{x},\mathbf{z},\mathbf{c},\mathbf{e}|\mathbf{y})\right]  \notag \\
&= \mathbb{E}_{q_\phi(\mathbf{x}|\mathbf{y})}\left[{\rm log}\,p_\theta(\mathbf{x},\mathbf{y})\right]+H(q_\phi(\mathbf{x}|\mathbf{y}))
\end{align}
The joint likelihood $p(\mathbf{x},\mathbf{y})$ is computed by marginalizing $\mathbf{z},\mathbf{c},\mathbf{e}$ from the forward variable $\alpha_t(\mathbf{z}_{t},\mathbf{c}_{t})$, and the approximate posterior distribution $q(\mathbf{x}|\mathbf{y})$ is computed by the amortized inference network.
The detailed training object of GRASS is described in App.~\ref{appendix: Detailed Optimization Objective of GRASS}.

\section{Experiments}

Most datasets for switching dynamical systems focus on scenarios with a single object switching dynamics, such as a one-dimensional bouncing ball, dubins path, a single dancer in Salsa Dancing from CMU MoCap ~\citep{dong2020collapsed}, and a 3 mode system \citep{ansari2021deep}.
While there are a few cases with multiple objects, these objects do not interact with one another. For instance, the dancing bees by~\citet{ansari2021deep} are considered a single ``super object'' comprising of all objects simultaneously. The two-dimensional reacher task by~\citet{dong2020collapsed} and neural populations by \citet{glaser2020recurrent} are similarly constructed.

By contrast, we focus on the generalized setting of having multiple objects that interact with one another, where interacting objects are considered simultaneously and depending on another with the objective of discovering dynamic modes and switching behaviours.
To evaluate the proposed methods and compare against baselines, we introduce two datasets for benchmarking, inspired by the single-object literature: the synthesized \emph{ODE-driven particle} dataset, and the \emph{Salsa Couple dancing} dataset. The code and datasets are available at \href{https://github.com/yongtuoliu/Graph-Switching-Dynamical-Systems}{https://github.com/yongtuoliu/Graph-Switching-Dynamical-Systems.}.

\paragraph{ODE-driven particle dataset.}
We introduce three Ordinary Differential Equation (ODE) systems as the three modes to generate time-evolving trajectories of particles, \emph{i.e.}, Lotka-Volterra, Spiral and Bouncing Ball ODE:
\begin{align}
    &{\rm Lotka\!-\!Volterra\!:}\,\,x^{\prime} = x - xy;\,\,y^{\prime} = -y + xy \\
    &{\rm Spiral\!:}\,\,x^{\prime} = -0.1x^3 + 2y^3;\,\,y^{\prime} = -2x^3 - 0.1y^3 \\
    &{\rm Bouncing\,\,Ball\!:}\,\,x^{\prime} = 0;\,\,y^{\prime} = 2\,({\rm or}\,y^{\prime} =-2)
\end{align}
To simulate trajectories, we draw balls with radius $r$, randomly initialized and driven by different ODEs on a squared 2d canvas of size 64*64.
Specifically, we consider three particle balls driven by three different ODE modes unless stated otherwise (\eg, in the experiments increasing the number of particles or the number of modes).
Numerical values of ODEs are mapped to the canvas.
For mode-switching interactions among objects, we switch the driven ODE modes of two objects when they collide in the canvas.
Each sample has 100 time steps, and with 10 frames per second.
{We follow the sample splitting proportion of synthesized datasets in REDSDS~\citep{ansari2021deep} (i.e. test data is around 5\% of training data) and create 4,928 samples for training, 191 samples for validation, and 204 samples for testing. Analyses on new splitting strategy (i.e. test data is around 10\% of training data) and larger dataset are in App.~\ref{appendix: New split and larger ODE}. A sample visualization of this dataset is shown in Fig.~\ref{ODE-figure}.

 \begin{figure}[t]
 \centering
 \setlength{\tabcolsep}{1pt}
 \includegraphics[width=.85\linewidth]{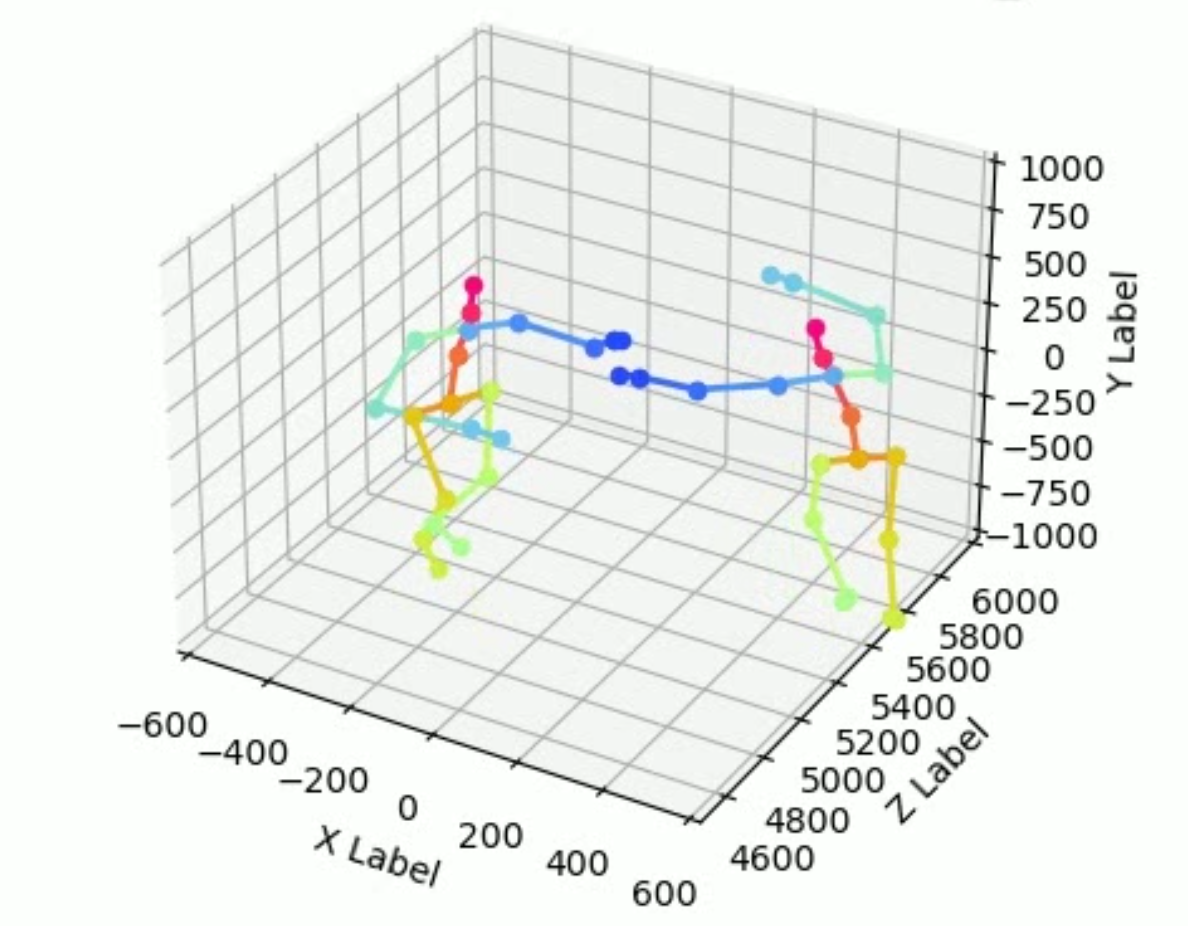}
  \vspace{-8pt}
 \caption{3D skeletons in Salsa Couple Dancing dataset.}
 \vspace{-8pt}
 \label{Salsa-Dancing-figure}
 \end{figure}

\paragraph{Salsa Couple dancing dataset.}
~\citet{dong2020collapsed} experiment with salsa dancing sequences, which, however, feature a single dancer only from CMU MoCap.
We collect 17 real-world Salsa dancing videos from the Internet, containing 8,672 frames. Among them, 3 videos are for testing and the remaining videos are for training. We extract 3D skeletons of dancers by a pretrained model~\citep{moon2019camera} and conduct temporal Gaussian smoothing afterward.
As \citet{dong2020collapsed}, we annotate four modes, i.e.,  ``moving forward'', ``moving  backward'', ``clockwise turning'', and ``counter-clockwise turning''. Each sample has 100 time steps with 5 frames per second.
We have 1,321 samples for training and 156 samples for testing.
The coordinates of 3D skeletal joints serve as input for each dancer, and the modes of each dancer at each time step are the output. In Fig.~\ref{Salsa-Dancing-figure} we show the 3D skeletons extracted from the videos.

\paragraph{Evaluation metrics.}
Following~\citet{dong2020collapsed,ansari2021deep}, we evaluate using frame-wise segmentation accuracy, \emph{i.e.} accuracy and $F_1$ after matching the labels using the Hungarian algorithm~\citep{kuhn1955hungarian}, Normalized Mutual Information (NMI) 
and Adjusted Rand Index (ARI) to measure similarity between two labellings.
We conduct each experiment for five random seeds and report the average performance and standard deviation of the results.

\paragraph{Baselines.}

We compare MOSDS and GRASS with three state-of-the-art methods: rSLDS~\citep{linderman2016recurrent}, SNLDS~\citep{dong2020collapsed}, and REDSDS~\citep{ansari2021deep}.
For our implementation, we use REDSDS~\citep{ansari2021deep} as the base for MOSDS and GRASS. 
We include in the comparisons GRASS-GT as an ``upper bound'' oracle method, for which we use the ground-truth graph edges rather to learn mode transition behaviours.

\subsection{ODE-driven Particle}

We summarize results for the ODE-drive particles in Table~\ref{ODE-driven Particle dataset table}.
We see that just by considering interactions between multiple objects with MOSDS, we achieve significant and consistent performance increases across all metrics.
When further using graphs to model the switching dynamics in our interacting system of objects, GRASS improves by more than 9-10\% over the previous state-of-the-art, REDSDS, across all metrics.
We also observe that GRASS performs similarly to GRASS-GT using ground truth edges, showcasing the accuracy of inferring the latent object-to-object interactions. In Fig.~\ref{ODE_vis_seg}, we show also the qualitative results of GRASS compared to REDSDS, which is the top performing baseline. GRASS discovers mode-switching behaviours between objects effectively and with fewer switching errors.

\begin{figure*}[t]
 \centering
 \setlength{\tabcolsep}{1pt}
 \includegraphics[width=.90\linewidth]{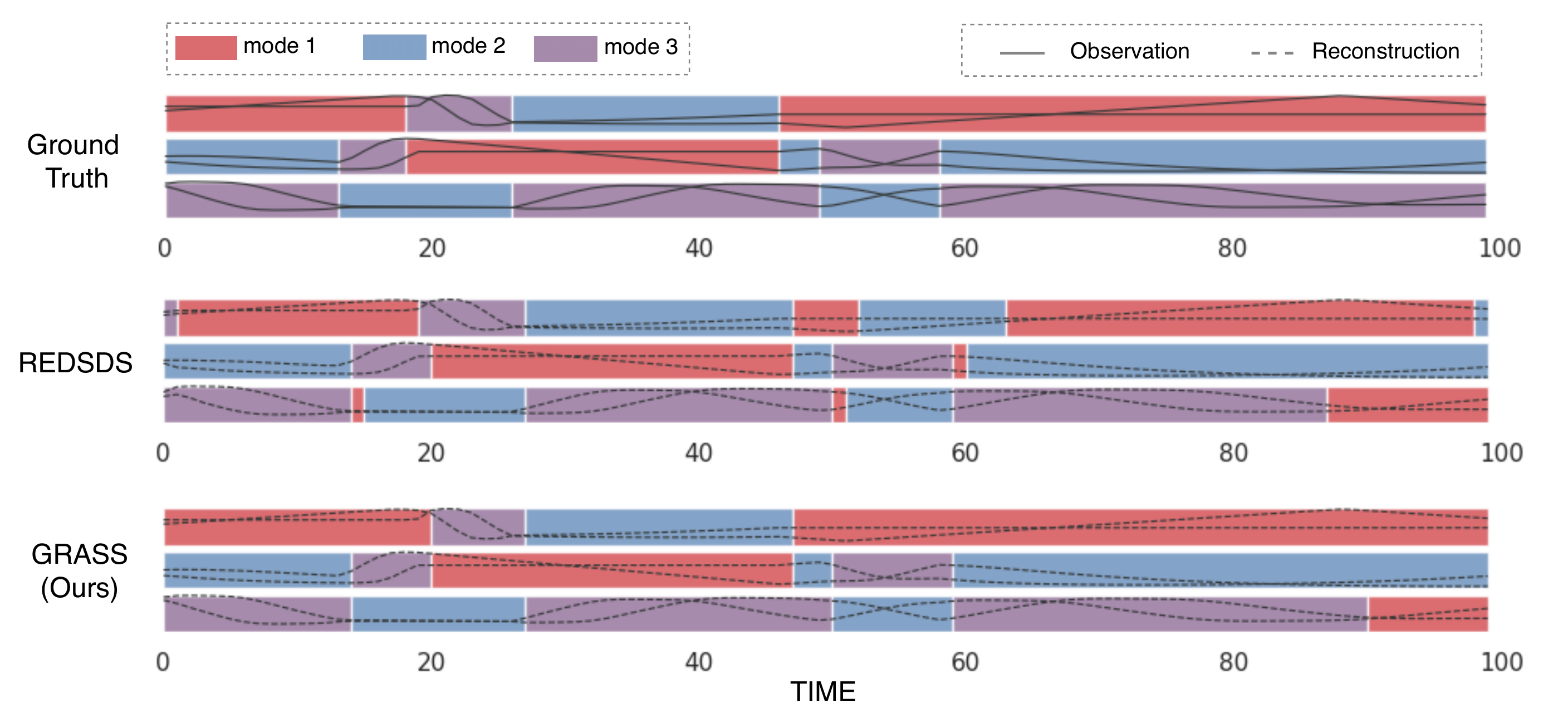}
 \vspace{-10pt}
 \caption{Qualitative results of our GRASS model compared to previous state-of-the-art method REDSDS~\citep{ansari2021deep}. Each row contains three sub-rows which denote the mode segmentation of multiple objects. We can see that with explicit interaction modeling by GRASS, mode-switching behaviors among objects are discovered effectively with fewer switching errors and better segmentation results.}
 \label{ODE_vis_seg}
 \end{figure*}

\begin{table}[htbp]
	\centering
	\scriptsize
	\tabcolsep=0.100cm
	\setlength\arrayrulewidth{1.0pt}
	\caption{Comparisons on ODE-driven Particle Dataset.}
	\vspace{4pt}
	\begin{tabular}{lcccc}
		\toprule
     Method & NMI $\uparrow$ & ARI $\uparrow$ & Accuracy $\uparrow$ & $F_1$ $\uparrow$ \\
		\midrule
        rSLDS  & 0.257$\pm$0.023 & 0.231$\pm$0.016 & 0.450$\pm$0.033 & 0.443$\pm$0.041 \cr
        SNLDS & 0.368$\pm$0.027 & 0.349$\pm$0.021 & 0.681$\pm$0.067 & 0.664$\pm$0.053 \cr
        REDSDS & 0.418$\pm$0.016 & 0.397$\pm$0.028 & 0.708$\pm$0.037 & 0.702$\pm$0.027\cr
        MOSDS (this paper) & 0.469$\pm$0.020 & 0.474$\pm$0.015 & 0.766$\pm$0.045 & 0.757$\pm$0.032 \cr
        GRASS (this paper) & \textbf{0.528$\pm$0.014} & \textbf{0.519$\pm$0.008} & \textbf{0.794$\pm$0.030} & \textbf{0.790$\pm$0.021}\cr
        \midrule
        GRASS-GT (Oracle) & {0.537$\pm$0.012} & {0.526$\pm$0.010} & {0.805$\pm$0.028} & {0.801$\pm$0.016}\cr
		\bottomrule
	\end{tabular}
	\label{ODE-driven Particle dataset table}
\end{table}

\subsection{Salsa Couple Dancing}

We summarize the results for Salsa Couple Dancing dataset in Table~\ref{Salsa Dancing dataset table}.
We observe similar findings in this real-world video dataset, as with the ODE-driven particles.
GRASS achieves significantly higher accuracy across all metrics, including REDSDS and our simpler method MOSDS.

\begin{table}[t]
	\centering
	\scriptsize
	\tabcolsep=0.100cm
	\setlength\arrayrulewidth{1.0pt}
	\caption{Comparisons on the Salsa Couple Dancing dataset.}
	\begin{tabular}{lcccc}
		\toprule
  Method & NMI $\uparrow$ & ARI $\uparrow$ & Accuracy $\uparrow$ & $F_1$ $\uparrow$ \\
		\midrule
        rSLDS & 0.118$\pm$0.028 & 0.102$\pm$0.043 & 0.373$\pm$0.066 & 0.360$\pm$0.053 \cr
        SNLDS & 0.145$\pm$0.047 & 0.133$\pm$0.031 & 0.420$\pm$0.113 & 0.413$\pm$0.096 \cr
        REDSDS & 0.156$\pm$0.032 & 0.152$\pm$0.036 & 0.504$\pm$0.052 & 0.467$\pm$0.074 \cr
        MOSDS (this paper)  & 0.162$\pm$0.053 & 0.165$\pm$0.072 & 0.537$\pm$0.091 & 0.508$\pm$0.063 \cr
        GRASS (this paper) & \textbf{0.174$\pm$0.031} & \textbf{0.176$\pm$0.043} & \textbf{0.569$\pm$0.065} & \textbf{0.524$\pm$0.046}\cr
		\bottomrule
	\end{tabular}
	\label{Salsa Dancing dataset table}
\end{table}

\subsection{Ablation experiments}
Due to limited space, we report the average performance in each table. Results with standard deviations are in App.~\ref{appendix: Ablation studies with error bars}.

\paragraph{Sensitivity to the number of interactions.}
We evaluate how sensitive is GRASS in the presence of an increasing number of interactions.
First, we extend the normal ODE-driven Particle dataset to more particles, \emph{i.e.} 3 particles, 5 particles, and 10 particles.
The number of interactions naturally increases with the number of particles in a space-constrained canvas.
For different numbers of particles, we count the average number of interactions per object per time series and they are 2.3 interactions for 3 particles, 6.1 for 5 particles, and 12.5 for 10 particles.
We present the results in Table~\ref{Number of Particles table}, where we conclude that GRASS is not adversely affected by an increasing number of objects and interactions.

\begin{table}[t]
	\centering
	\scriptsize
	\tabcolsep=0.100cm
	\setlength\arrayrulewidth{1.0pt}
	\caption{Analyses on different numbers of objects on ODE-driven Particle dataset, while \emph{increasing} the average number of interactions per object per time series, i.e, 2.3 interactions for 3 particles, 6.1 for 5, and 12.5 for 10. */* denotes NMI / $F_1$.}
	\vspace{4pt}
	\begin{tabular}{lccc}
		\toprule
  Number of Particles & 3 & 5 & 10 \\
		\midrule
        rSLDS & 0.257 / 0.443 & 0.252 / 0.437 & 0.246 / 0.430 \cr
        SNLDS & 0.368 / 0.664 & 0.361 / 0.656 & 0.354 / 0.651 \cr
        REDSDS & 0.418 / 0.701 & 0.411 / 0.692 & 0.405 / 0.687 \cr
        MOSDS (this paper) & 0.469 / 0.757  & 0.461 / 0.752  & 0.456 /  
 0.748 \cr
        GRASS (this paper) & \textbf{0.528} / \textbf{0.790} & \textbf{0.524} / \textbf{0.784} & \textbf{0.519} / \textbf{0.781} \cr
		\bottomrule
	\end{tabular}
	\label{Number of Particles table}
\end{table}

\paragraph{Sensitivity to the number of objects.}
We further test increasing the number of objects, while fixing the number of interactions. We achieve this by controlling the sizes of objects, as with smaller balls we have fewer collisions (and thus interactions).
We roughly fix the number of interactions per object per time series to be 2.3 and change the number of objects to 3, 5, and 10 as in the previous trial.
We present results in Table~\ref{Number of interactions table}.
GRASS is robust to different numbers of objects, no matter whether we fix the number of interactions.
\begin{table}[t]
	\centering
	\scriptsize
	\tabcolsep=0.100cm
	\setlength\arrayrulewidth{1.0pt}
	\caption{Analyses on different numbers of objects on ODE-driven Particle, while \emph{fixing} the average number of interactions per object per time series, i.e, 2.3 interactions. */* denotes NMI / $F_1$.}
	\vspace{4pt}
	\begin{tabular}{lccc}
		\toprule
  Number of Particles & 3 & 5 & 10 \\
		\midrule
        rSLDS & 0.257 / 0.443 & 0.262 / 0.444 & 0.253 / 0.437 \cr
        SNLDS & 0.368 / 0.664 & 0.365 / 0.666 & 0.362 / 0.659 \cr
        REDSDS & 0.418 / 0.701 & 0.423 / 0.706 & 0.413 / 0.694\cr
        MOSDS (this paper) & 0.469 / 0.757 & 0.471 / 0.763 & 0.464 / 0.754 \cr
        GRASS (this paper) & \textbf{0.528} / \textbf{0.790} & \textbf{0.530} / \textbf{0.792} & \textbf{0.524} / \textbf{0.786}\cr
		\bottomrule
	\end{tabular}
	\label{Number of interactions table}
\end{table}

\paragraph{Sensitivity to absence of interactions.}
GRASS is built for systems of multiple objects that interact with one another.
We test whether the method generalizes even in the case when the objects are independent and do not interact, as with single-object Switching Dynamical Systems.
We create a dataset with three particles driven by three different ODEs, and set them so that they do not interact with each other.
We present results in Table~\ref{no interaction table}.
In the presence of interactions, GRASS is considerably more accurate than REDSDS, while in the absence of interactions, it scores comparably. 
In this case MOSDS observes a higher drop in accuracy.
The reason is that with its dynamic graph, GRASS can still predict correctly that there exist no interaction edges between objects, while MOSDS always assumes all objects interact.

\begin{table}[t]
	\centering
	\scriptsize
	\tabcolsep=0.100cm
	\setlength\arrayrulewidth{1.0pt}
	\caption{Analyses of robustness to datasets without interactions on ODE-driven Particle dataset. */* denotes NMI / $F_1$.}
	\vspace{4pt}
	\begin{tabular}{lcc}
		\toprule
  Method & dataset w/ interaction & dataset w/o interaction  \\
		\midrule
        rSLDS & 0.257 / 0.443 &  0.471 / 0.686 \cr
        SNLDS & 0.368 / 0.664 & 0.534 / 0.772 \cr
        REDSDS & 0.418 / 0.701 & \textbf{0.579} / \textbf{0.838} \cr
        MOSDS (this paper) & 0.469 / 0.757  & 0.563/ 0.817 \cr
        GRASS (this paper) & \textbf{0.528} / \textbf{0.790} & 0.573 / 0.826 \cr
		\bottomrule
	\end{tabular}
	\label{no interaction table}
\end{table}

\paragraph{Sensitivity to number of dynamic modes.}
Like previous methods~\citep{linderman2016recurrent, dong2020collapsed, ansari2021deep}, GRASS requires a predefined maximum number of modes.
We test its robustness to different maximum numbers of modes, that is 3, 5, and 10, while the true number of modes is 3. 
We present results in Table~\ref{Number of Modes table}.
We observe that GRASS is impervious to this misspecification, which suggests that we can set a large number of possible modes and GRASS will still use only those needed.

\begin{table}[t]
	\centering
	\scriptsize
	\tabcolsep=0.100cm
	\setlength\arrayrulewidth{1.0pt}
	\caption{Analyses on robustness to different maximal numbers of predefined modes. */* denotes NMI / $F_1$.}
	\vspace{4pt}
	\begin{tabular}{lccc}
		\toprule
  Number of Modes & 3 & 5 & 10 \\
		\midrule
        rSLDS & 0.257 / 0.443 & 0.253 / 0.438 & 0.248 / 0.436 \cr
        SNLDS & 0.368 / 0.664 & 0.365 / 0.661 & 0.362 / 0.657 \cr
        REDSDS & 0.418 / 0.701 & 0.415 / 0.696 & 0.413 / 0.694 \cr
        MOSDS (this paper) & 0.469 / 0.757  & 0.466 / 0.759  & 0.462 / 0.754\cr
        GRASS (this paper) &\textbf{0.528} / \textbf{0.790} & \textbf{0.532} / \textbf{0.794} & \textbf{0.527} / \textbf{0.784} \cr 
		\bottomrule
	\end{tabular}
	\label{Number of Modes table}
\end{table}

\section{Related Work}

Switching Linear Dynamical Systems (SLDS) \citep{ackerson1970state,ghahramani2000variational,oh2005variational} 
introduce both discrete states to represent motion modes and continuous states to characterize motion dynamics of each mode, but assume linear state transitions.
Switching Non-linear Dynamical Systems, implemented by neural networks, extend these methods to the nonlinear case, providing a better expressiveness of complex system dynamics.
Among them, SNLDS \citep{dong2020collapsed} and REDSDS \citep{ansari2021deep} are two representative methods that can consistently outperform their linear counterparts. 
While effective, previous methods and datasets are usually limited to single-object scenarios where only one object exist. When multiple objects exist, objects are processed independently or considered as one single super-object with a single mode.
For example, in \citep{glaser2020recurrent}, multiple neural populations exist in the brain, while the only mode behaviours of the whole brain only are modelling and discovered.
By contrast, in this paper we focus on the general setting where our systems comprise multiple objects interacting and changing their behaviour accordingly.

Graph Neural Networks are the \emph{de facto} choice for learning relational representations over graphs.
Recently, there are some methods focusing on neural relational inference \citep{kipf2018neural,Graber_2020_CVPR, kofinas2021roto} over temporal sequences, whose dynamics are encoded by continuous latent states.
These methods focus on systems with multiple objects, whose dynamics, however, do not change of time and, therefore, are not a good fit for discovering mode-switching behaviours over time.
In this work, we start from the framework of Switching Dynamical Systems, and integrate them within a graph neural network formalism.
In particular, we extend neural relational graphs and relational inference \citep{kipf2018neural, Graber_2020_CVPR} to incorporate latent interaction variables, one per pair of objects, and model the potential dynamic interactions between objects.
The proposed Graph Switching Dynamical Systems can thus handle systems with increased complexity with a significantly better accuracy. This is true even in the presence of sparse interactions in both space and time, which cause sudden and complex dynamic mode switches.

\section{Conclusion and Future work}

We investigate the setting of \emph{interacting objects} switching dynamical systems, when objects interact with each other and influence each other's modes.
We propose a graph-based approach for these systems, GRAph Switching dynamical Systems (GRASS), in which we use a dynamic graph to model interactions and mode-switching behaviors between objects.
We also introduce two datasets, \emph{i.e.} a synthesized ODE-driven Particle dataset and a real-world Salsa Couple dancing dataset.
Experiments show that GRASS improves considerably the state-of-the-art. Future work includes exploring learning switching dynamical systems with multiple objects directly from videos.

\section*{Acknowledgements}
This work is financially supported by NWO TIMING VI.Vidi.193.129.
We also thank SURF for the support in using the National Supercomputer Snellius.

\newpage

\bibliography{icml_paper}
\bibliographystyle{icml2023}

\newpage
\appendix
\onecolumn
{\Large \textbf{Appendix}}
\section{More details of GRASS model}

\subsection{Inference Algorithm of GRASS}
\label{appendix: Inference Algorithm}
The inference algorithm of GRASS is in Alg.~\ref{optimization procedure}. 
As inputs, we have a time series $\mathbf{y}_{1:T}$ and an interaction edge prior distribution $p(\mathbf{e}_{1:T})$. First, we initialize distributions of continuous state and discrete mode variables as $p(\mathbf{x}_1)$ and $p(\mathbf{z}_1)$. Besides, the range of discrete count variable is initialized as $\{d_{min}, ..., d_{max}\}$. For each time step $t$ in the time series, the continuous state and discrete edge are first inferred by posterior approximation, i.e. $\tilde{\mathbf{x}}_{t}\sim q_{\phi_x}\!(\mathbf{x}_{t}|\mathbf{y}_{1:T})$ and $\tilde{\mathbf{e}_t}\sim q_{\phi_e}\!(\mathbf{e}_t|\tilde{\mathbf{x}}_{t})$. Then we calculate continuous state and discrete mode transition probabilities, i.e. $p_{\theta_{xtr}}(\mathbf{x}_t|\tilde{\mathbf{x}}_{t-1}, \mathbf{z}_{t})$ and $p_{\theta_{ztr}}(\mathbf{z}_{t}^{1:N}|\mathbf{z}_{t-1}^{1:N}, \tilde{\mathbf{x}}_{t-1}^{1:N}, \mathbf{c}_{t}^{1:N})$, which are used for exact inference of discrete mode and count by calculating forward and backward variables $\alpha_t(\mathbf{z}_{t},\mathbf{c}_{t})$ and $\beta_t(\mathbf{z}_{t},\mathbf{c}_{t})$ in Forward-and-Backward algorithm. Besides, two consistency losses are introduced by calculating the loglikelihood between $\tilde{\mathbf{x}_{t}}$ and $\hat{\mathbf{x}_{t}}$, $\tilde{\mathbf{y}_{t}}$ and $\mathbf{y}_{t}$. We finally derive the ELBO optimization objective to optimize the parameters of networks. Details of the derivatives of ELBO are in Section \ref{appendix: Detailed Optimization Objective of GRASS}. An illustration of the inference stage is in Fig.~\ref{architecture}. Besides, the overall generative model and inference stages of GRASS which factorize objects are detailed in Fig.~\ref{factorized-figure}. 

\begin{algorithm}
\caption{Inference algorithm for GRASS.}
\label{optimization procedure}
\LinesNumbered
\KwIn{Time series $\mathbf{y}_{1:T}$, interaction edge prior distribution $p(\mathbf{e}_{1:T})$}
\KwOut{Learned parameters $\phi$ and $\theta$.}

Initialize prior continuous state and discrete mode distributions as $p(\mathbf{x}_{1})$, $p(\mathbf{z}_{1})$; Initialize the range of discrete count variable $\{d_{min}, ..., d_{max}\}$ ;

\For{t \emph{in} $[1, \dots, T]$}{
\tcp{State Inference}

Infer continuous state $\tilde{\mathbf{x}}_{t}\sim q_{\phi_x}\!(\mathbf{x}_{t}|\mathbf{y}_{1:T})$;

Infer discrete edge $\tilde{\mathbf{e}_t}\sim q_{\phi_e}\!(\mathbf{e}_t|\tilde{\mathbf{x}}_{t})$;

\tcp{Calculate continuous state transition}
Calculate continuous state transition $\hat{\mathbf{x}}_{t} \sim p_{\theta_{xtr}}(\mathbf{x}_t|\tilde{\mathbf{x}}_{t-1}, \mathbf{z}_{t})$;

\tcp{Calculate discrete mode transition}

\For{n, m $\in [1, \dots, N]$}{
Calculate interaction weights $w_t^{m \rightarrow n} = \sum_{l=2}^{L+1} \tilde{e}^{m\rightarrow n}_{t, l}$;

Calculate $\tilde{\mathbf{x}}^{m, n}_{t-1} = \sum_{l} \tilde{e}^{m\rightarrow n}_{t, l}\cdot f_e^l([\tilde{\mathbf{x}}_{t-1}^{m}, \tilde{\mathbf{x}}_{t-1}^{n}])$;

Calculate $p_{\theta_{ztr}}(z_{t}^{n}|z_{t-1}^{m}, \tilde{\mathbf{x}}_{t-1}^{m,n}, c_{t}^{n}, \tilde{e}_t^{m\rightarrow n})$;

}

Calculate discrete mode transition 
 $p_{\theta_{ztr}}(\mathbf{z}_{t}^{1:N}|\mathbf{z}_{t-1}^{1:N}, \tilde{\mathbf{x}}_{t-1}^{1:N}, \mathbf{c}_{t}^{1:N}) =  \prod_{n=1}^{N}\!\sum_{m=1}^{N}\!\!w_t^{m\rightarrow n}\,p_{\theta_{ztr}}(z_{t}^{n}|z_{t-1}^{m}, \tilde{\mathbf{x}}_{t-1}^{m,n}, c_{t}^{n}, \tilde{e}_t^{m\rightarrow n})$;

\tcp{Reconstruct input}
Emit reconstructed input $\tilde{\mathbf{y}_{t}}\sim p_{\theta_y}(\mathbf{y}_{t}|\tilde{\mathbf{x}_{t}})$;

\tcp{Log-likelihood Calculation}

Calculate LogLikelihood($\tilde{\mathbf{y}_{t}}$, $\mathbf{y}_{t}$);

Calculate LogLikelihood($\tilde{\mathbf{x}_{t}}$, $\hat{\mathbf{x}_{t}}$);

\tcp{Exact inference of discrete mode and count}

Calculate Forward algorithm variable: $\alpha_t(\mathbf{z}_{t},\mathbf{c}_{t}) = p(\mathbf{y}_{1:t},\tilde{\mathbf{x}}_{1:t},\tilde{\mathbf{e}}_{1:t}, \mathbf{z}_{1:t},\mathbf{c}_{1:t})$

Calculate Backward algorithm variable:
$\beta_t(\mathbf{z}_{t},\mathbf{c}_{t}) = p(\mathbf{y}_{t+1:T},\tilde{\mathbf{x}}_{t+1:T}\,|\,\tilde{\mathbf{x}}_{t},\tilde{\mathbf{e}}_{t}, \mathbf{z}_{t},\mathbf{c}_{t})$;
}
\tcp{ELBO optimization}

$\mathrm{argmax}_{\phi, \theta} \ {\rm log}\,p_\theta(\mathbf{y})\!-\!D_{K\!L}\left[q_\phi(\mathbf{x},\mathbf{z},\mathbf{c},\mathbf{e}|\mathbf{y})\,\|\,p_\theta(\mathbf{x},\mathbf{z},\mathbf{c},\mathbf{e}|\mathbf{y})\right]$

\end{algorithm}

\begin{figure*}
 \centering
 \setlength{\tabcolsep}{1pt}
 \includegraphics[width=.90\linewidth]{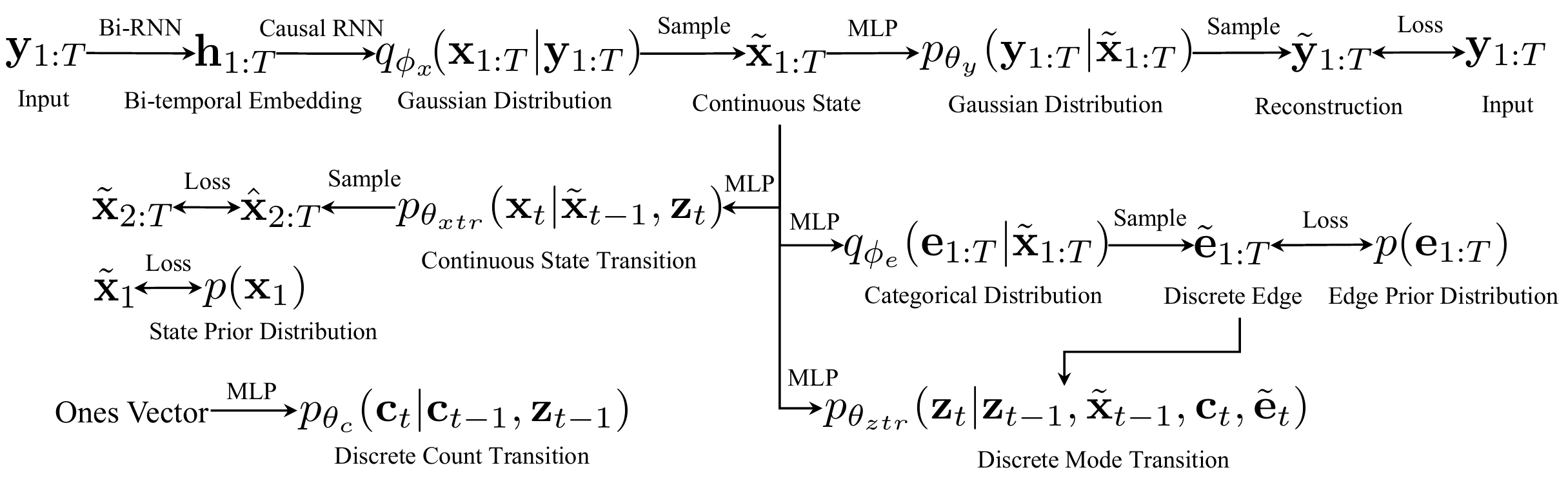}
  \vspace{-4pt}
 \caption{Illustration of inference algorithm of Graph Switching Dynamical Systems. After the approximate inference of continuous state $\tilde{\mathbf{x}}_{1:T}$ and discrete edge $\tilde{\mathbf{e}}_{1:T}$, we further calculate continuous state transition probability $p_{\theta_{xtr}}(\mathbf{x}_t|\tilde{\mathbf{x}}_{t-1}, \mathbf{z}_t)$, discrete mode transition probability $p_{\theta_{ztr}}(\mathbf{z}_t|\mathbf{z}_{t-1}, \tilde{\mathbf{x}}_{t-1}, \mathbf{c}_t, \tilde{\mathbf{e}}_t)$, and discrete count transition probability $p_{\theta_{c}}(\mathbf{c}_t|\mathbf{c}_{t-1}, \mathbf{z}_{t-1})$, which are utilized by the forward and backward algorithm to conduct exact inference of discrete mode $\mathbf{z}_{1:T}$ and count $\mathbf{c}_{1:T}$ to finally derive ELBO optimization objective.}
 \vspace{-2pt}
 \label{architecture}
 \end{figure*}

 \begin{figure*}
 \centering
 \setlength{\tabcolsep}{1pt}
 \includegraphics[width=.90\linewidth]{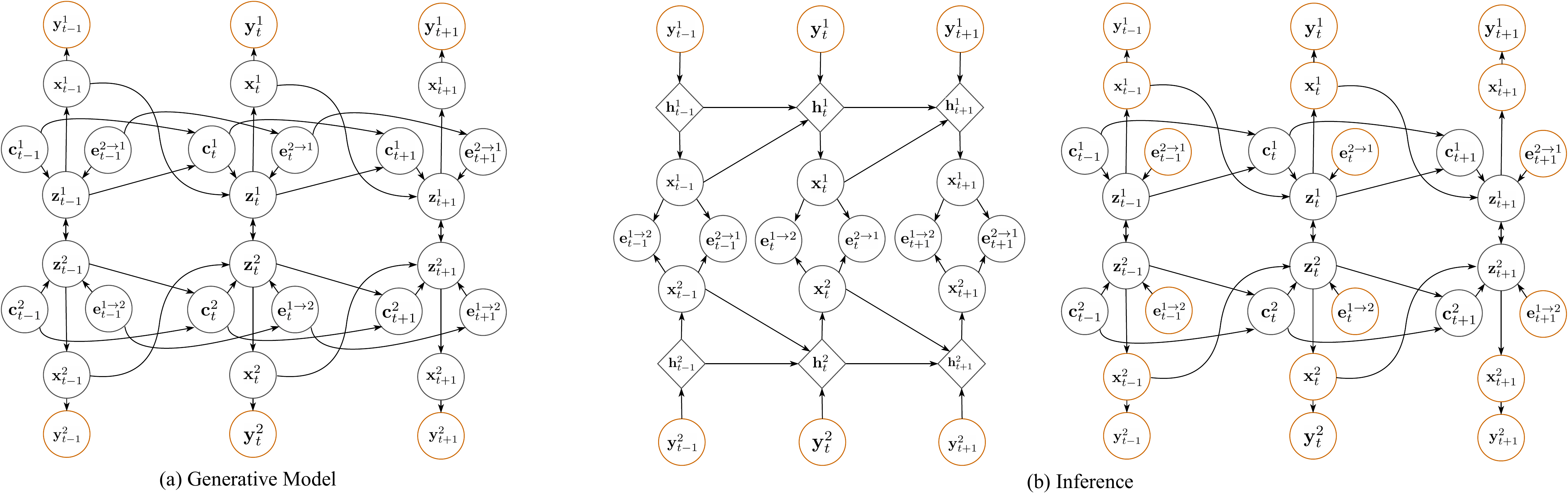}
  \vspace{-4pt}
 \caption{(a) Generative model of GRASS. (b) Left: Amortized approximate inference for the continuous states (e.g. $\mathbf{x}_{t}^{1}$ and $\mathbf{x}_{t}^{2}$) and discrete edge variable (e.g. $\mathbf{e}_{t}^{1\rightarrow2}$ and $\mathbf{e}_{t}^{2\rightarrow1}$) by inference networks. Temporal dependence is modeled by an intermediate latent embedding (e.g. $\mathbf{h}_{t}^{1}$ and $\mathbf{h}_{t}^{2}$) which is given by directional RNNs. Right: Exact inference of discrete mode (e.g. $\mathbf{z}_{t}^{1}$ and $\mathbf{z}_{t}^{2}$) and count variables  and (e.g. $\mathbf{c}_{t}^{1}$ and $\mathbf{c}_{t}^{2}$) based on the approximate pseudo-observations (e.g. $\mathbf{x}_{t}^{1}$ and $\mathbf{x}_{t}^{2}$) and pseudo-interactions (e.g. $\mathbf{e}_{t}^{1\rightarrow2}$ and $\mathbf{e}_{t}^{2\rightarrow1}$). Orange circles denote observations or approximate pseudo-observations. Here, we assume there exist two objects in the scenario.}
 \vspace{-2pt}
 \label{factorized-figure}
 \end{figure*}

 \subsection{Implementation Details}
 \label{appendix: Implementation Details}
 In the following, we show the network details as well as embedding dimensions. $\mathtt{biGRU\,[a]}$ denotes a bidirectional GRU with a single-layer of $\mathtt{a}$ hidden units.
$\mathtt{MLP\,[b]}$ denotes a single-layer MLP with $\mathtt{b}$ hidden units and ReLU non-linearity.
$\mathtt{RNN\,[c]}$ denotes a single-layer RNN with $\mathtt{c}$ hidden units. Inference networks for continuous state $\mathbf{x}$: $\mathtt{biGRU\,[4]}$, $\mathtt{RNN\,[16]}$, and $\mathtt{MLP\,[8]}$;
Inference networks for discrete edge $\mathbf{e}$: $\mathtt{MLP\,[128]}$ (i.e. $f_{\phi_z}^{\rm emb}$), $\mathtt{MLP\,[128]}$ (i.e. $f_{\phi_z}^{e,1}$), $\mathtt{MLP\,[128]}$ (i.e. $f_{\phi_z}^{v,1}$), and $\mathtt{MLP\,[2]}$ for ODE-driven particle dataset or $\mathtt{MLP\,[5]}$ for Salsa-couple dancing dataset (i.e. $f_{\phi_z}^{e,2}$);
Continuous transition network: $\mathtt{MLP\,[8]}$ (i.e. $p(\mathbf{x}_t^{n}|\mathbf{x}_{t-1}^{n}, z_{t}^{n})$);
Discrete transition network: $\mathtt{MLP\,[2^2]}$ for ODE-driven particle dataset or $\mathtt{MLP\,[4^4]}$ for ODE-driven particle dataset (i.e. $p(z_{t}^{n}|z_{t-1}^{m}, \mathbf{x}_{t-1}^{m,n}, c_{t}^{n}, e_t^{m\rightarrow n})$);
Emission network: $\mathtt{MLP\,[2]}$ for ODE-driven particle dataset or $\mathtt{MLP\,[45]}$ for ODE-driven particle dataset (i.e. $p(\mathbf{y}_t^{n}|\mathbf{x}_t^{n})$).

We train both datasets with a fixed batch size of 20 for 60,000 training steps. We use the Adam optimizer with $10^{-5}$ weight-decay and clip gradients norm to 10. The learning rate is warmed up linearly from $5\times10^{-5}$ to $2\times10^{-4}$ for the first 2,000 steps, and then decays following a cosine manner with a rate of 0.99. Each experiment is running on one Nvidia GeForce RTX 3090 GPU.

\subsection{Detailed Optimization Objective of GRASS}
\label{appendix: Detailed Optimization Objective of GRASS}
\subsubsection{Derivation of ELBO}
The evidence lower bound objective (ELBO) of Graph Switching Dynamical System (GRASS) is defined as follows. For brevity, $\mathbf{x}$, $\mathbf{y}$, $\mathbf{z}$, $\mathbf{c}$, and $\mathbf{e}$ represents $\mathbf{x}_{1:T}^{1:N}$, $\mathbf{y}_{1:T}^{1:N}$, $\mathbf{z}_{1:T}^{1:N}$, $\mathbf{c}_{1:T}^{1:N}$, and $\mathbf{e}_{1:T}^{1:N^2}$ respectively. $N$ is the number of objects. $T$ is the number of timestamps.
\begin{align*}
ELBO &= {\rm log}\,p_\theta(\mathbf{y})\!-\!D_{K\!L}\left[q_\phi(\mathbf{x},\mathbf{z},\mathbf{c},\mathbf{e}|\mathbf{y})\,\|\,p_\theta(\mathbf{x},\mathbf{z},\mathbf{c},\mathbf{e}|\mathbf{y})\right] \\
&= \int q_\phi(\mathbf{x},\mathbf{z},\mathbf{c},\mathbf{e}|\mathbf{y}) \, {\rm log}\,p_\theta(\mathbf{y}) \, d(\mathbf{x},\mathbf{z},\mathbf{c},\mathbf{e}) - \int q_\phi(\mathbf{x},\mathbf{z},\mathbf{c},\mathbf{e}|\mathbf{y}) \, {\rm log}\,\frac{q_\phi(\mathbf{x},\mathbf{z},\mathbf{c},\mathbf{e}|\mathbf{y})}{p_\theta(\mathbf{x},\mathbf{z},\mathbf{c},\mathbf{e}|\mathbf{y})} \, d(\mathbf{x},\mathbf{z},\mathbf{c},\mathbf{e}) \\
&= \int q_\phi(\mathbf{x},\mathbf{z},\mathbf{c},\mathbf{e}|\mathbf{y})\left[{\rm log}\,p_\theta(\mathbf{x},\mathbf{z},\mathbf{c},\mathbf{e},\mathbf{y})-{\rm log}\,q_\phi(\mathbf{x},\mathbf{z},\mathbf{c},\mathbf{e}|\mathbf{y})\right] \, d(\mathbf{x},\mathbf{z},\mathbf{c},\mathbf{e}) \\
&= \mathbb{E}_{q_\phi(\mathbf{x},\mathbf{z},\mathbf{c},\mathbf{e}|\mathbf{y})}\left[{\rm log}\,p_\theta(\mathbf{x},\mathbf{z},\mathbf{c},\mathbf{e},\mathbf{y})-{\rm log}\,q_\phi(\mathbf{x},\mathbf{z},\mathbf{c},\mathbf{e}|\mathbf{y})\right] \\
&= \mathbb{E}_{q_\phi(\mathbf{x}|\mathbf{y})q_\phi(\mathbf{e}|\mathbf{x})p_\theta(\mathbf{z},\mathbf{c}|\mathbf{x},\mathbf{y},\mathbf{e})}\left[{\rm log}\,p_\theta(\mathbf{x},\mathbf{y})q_\phi(\mathbf{e}|\mathbf{x})p_\theta(\mathbf{z},\mathbf{c}|\mathbf{x},\mathbf{y},\mathbf{e})-{\rm log}\,q_\phi(\mathbf{x}|\mathbf{y})q_\phi(\mathbf{e}|\mathbf{x})p_\theta(\mathbf{z},\mathbf{c}|\mathbf{x},\mathbf{y},\mathbf{e})\right] \\
&= \mathbb{E}_{q_\phi(\mathbf{x}|\mathbf{y})q_\phi(\mathbf{e}|\mathbf{x})p_\theta(\mathbf{z},\mathbf{c}|\mathbf{x},\mathbf{y},\mathbf{e})}\left[{\rm log}\,p_\theta(\mathbf{x},\mathbf{y})-{\rm log}\,q_\phi(\mathbf{x}|\mathbf{y})\right] \\
&= 
\mathbb{E}_{q_\phi(\mathbf{x}|\mathbf{y})}\left[{\rm log}\,p_\theta(\mathbf{x},\mathbf{y})-{\rm log}\,q_\phi(\mathbf{x}|\mathbf{y})\right]\\
&= 
\mathbb{E}_{q_\phi(\mathbf{x}|\mathbf{y})}\left[{\rm log}\,p_\theta(\mathbf{x},\mathbf{y})\right]+H(q_\phi(\mathbf{x}|\mathbf{y})),
\end{align*}
where the first term is a model likelihood, and the second term is conditional entropy for variational posterior of continuous latent state $\mathbf{x}$. With the proper assumption of conditional independence of continuous latent states among objects, the conditional entropy is expanded through space and time as:
\begin{align*}
H(q_\phi(\mathbf{x}|\mathbf{y})) &= H(q_\phi(\mathbf{x}_{1:T}^{1:N}|\mathbf{y}_{1:T}^{1:N})) \\
& = H\left(\prod_{n=1}^{N}q_\phi(\mathbf{x}_{1:T}^{n}|\mathbf{y}_{1:T}^{n})\right) \\
& = \sum_{n=1}^{N}H(q_\phi(\mathbf{x}_{1:T}^{n}|\mathbf{y}_{1:T}^{n})) \\
& = \sum_{n=1}^{N}H\left[(q_\phi(\mathbf{x}_{1}^{n}|\mathbf{y}_{1}^{n})\prod_{t+2}^Tq_\phi(\mathbf{x}_{t}^{n}|\Tilde{\mathbf{x}}_{1:t-1}^{n},\mathbf{y}_{t}^{n}))\right]\\
& = \sum_{n=1}^{N}\left[H(q_\phi(\mathbf{x}_{1}^{n}|\mathbf{y}_{1}^{n}))+\sum_{t=2}^{T}H(q_\phi(\mathbf{x}_{t}^{n}|\Tilde{\mathbf{x}}_{1:t-1}^{n},\mathbf{y}_{t}^{n}))\right]
\end{align*}
where $\Tilde{\mathbf{x}}_{1:t-1}^{n}$ contains $\Tilde{\mathbf{x}}_{1}^{n}$, $\Tilde{\mathbf{x}}_{2}^{n}$, ..., $\Tilde{\mathbf{x}}_{t-1}^{n}$, in which $\Tilde{\mathbf{x}}_{t-1}^{n} \sim q_\phi(\mathbf{x}_{t-1}^{n}|\Tilde{\mathbf{x}}_{1:t-2}^{n},\mathbf{y}_{t-1}^{n})$ is sampled from the variational posterior distribution. In practice, we utilize causal RNN to model the temporal dependence.

\subsubsection{Training of ELBO}
For training, we utilize mini-batch stochastic gradient descent algorithm. The gradients with respect to $\theta$ or $\phi$ in ELBO are calculated as:
\begin{align*}
 \nabla_\theta ELBO &= \nabla_\theta \left[\mathbb{E}_{q_\phi(\mathbf{x}|\mathbf{y})}{\rm log}\,p_\theta(\mathbf{x},\mathbf{y})\right] = \mathbb{E}_{q_\phi(\mathbf{x}|\mathbf{y})}\nabla_\theta{\rm log}\,p_\theta(\mathbf{x},\mathbf{y}), \\
\nabla_\phi ELBO &= \nabla_\phi \left[\mathbb{E}_{q_\phi(\mathbf{x}|\mathbf{y})}{\rm log}\,p_\theta(\mathbf{x},\mathbf{y})+H(q_\phi(\mathbf{x}|\mathbf{y}))\right] \\
& = \nabla_\phi \left[\mathbb{E}_{q_\phi(\mathbf{x}|\mathbf{y})}{\rm log}\,p_\theta(\mathbf{x},\mathbf{y})\right] + \nabla_\phi H(q_\phi(\mathbf{x}|\mathbf{y}))\\
& = \mathbb{E}_{\,\mathbf{\epsilon} \sim \mathcal{N}}\left[\nabla_\phi{\rm log}\,p_\theta(\mathbf{x},\mathbf{y}_\phi(\mathbf{x}, \mathbf{\epsilon}))\right] + \nabla_\phi H(q_\phi(\mathbf{x}|\mathbf{y})),
\end{align*}
where we use the reparameterization trick~\citep{kingma2013auto} to calculate gradient of $\nabla_\phi \left[\mathbb{E}_{q_\phi(\mathbf{x}|\mathbf{y})}{\rm log}\,p_\theta(\mathbf{x},\mathbf{y})\right]$.

Analyzing both $\nabla_\theta {\rm ELBO}$ and $\nabla_\phi {\rm ELBO}$, the challenging part is $\nabla_{\theta, \phi}{\rm log}\,p_\theta(\mathbf{x},\mathbf{y})$. Following \citep{ansari2021deep}, the derivative of the log-joint likelihood $\nabla{\rm log}\,p(\mathbf{x},\mathbf{y})$ is calculated as:
\begin{align*}
\nabla{\rm log}\,p(\mathbf{x},\mathbf{y}) 
&= \mathbb{E}_{p(\mathbf{z},\mathbf{c},\mathbf{e}|\mathbf{x},\mathbf{y})}\left[\nabla{\rm log}\,p(\mathbf{x},\mathbf{y})\right] \\
&= \mathbb{E}_{p(\mathbf{z},\mathbf{c},\mathbf{e}|\mathbf{x},\mathbf{y})}\left[\nabla{\rm log}\,p(\mathbf{x},\mathbf{y},\mathbf{z},\mathbf{c},\mathbf{e})\right] - \mathbb{E}_{p(\mathbf{z},\mathbf{c},\mathbf{e}|\mathbf{x},\mathbf{y})}\left[\nabla{\rm log}\,p(\mathbf{z},\mathbf{c},\mathbf{e}|\mathbf{x},\mathbf{y})\right] \\
&= \mathbb{E}_{p(\mathbf{z},\mathbf{c},\mathbf{e}|\mathbf{x},\mathbf{y})}\left[\nabla{\rm log}\,p(\mathbf{x},\mathbf{y},\mathbf{z},\mathbf{c},\mathbf{e})\right],
\end{align*}
where $\mathbb{E}_{p(\mathbf{z},\mathbf{c},\mathbf{e}|\mathbf{x},\mathbf{y})}\left[\nabla{\rm log}\,p(\mathbf{z},\mathbf{c},\mathbf{e}|\mathbf{x},\mathbf{y})\right]$ is calculated as:
\begin{align*}
\mathbb{E}_{p(\mathbf{z},\mathbf{c},\mathbf{e}|\mathbf{x},\mathbf{y})}\left[\nabla{\rm log}\,p(\mathbf{z},\mathbf{c},\mathbf{e}|\mathbf{x},\mathbf{y})\right] &= \int p(\mathbf{z},\mathbf{c},\mathbf{e}|\mathbf{x},\mathbf{y})\frac{\nabla{\rm log}\,p(\mathbf{z},\mathbf{c},\mathbf{e}|\mathbf{x},\mathbf{y})}{p(\mathbf{z},\mathbf{c},\mathbf{e}|\mathbf{x},\mathbf{y})}d(\mathbf{z},\mathbf{c},\mathbf{e}) \\
&= \nabla\int{\rm log}\,p(\mathbf{z},\mathbf{c},\mathbf{e}|\mathbf{x},\mathbf{y})d(\mathbf{z},\mathbf{c},\mathbf{e}) = \nabla 1 = 0,
\end{align*}
With Markovian property, we rewrite $\nabla{\rm log}\,p(\mathbf{x},\mathbf{y},\mathbf{z},\mathbf{c},\mathbf{e})$ as:
\begin{align*}
    &\nabla{\rm log}\,p(\mathbf{x},\mathbf{y},\mathbf{z},\mathbf{c},\mathbf{e}) \\
    &= \nabla\,{\rm log}\,p(\mathbf{x}_{1:T}^{1:N},\mathbf{y}_{1:T}^{1:N},\mathbf{z}_{1:T}^{1:N},\mathbf{c}_{1:T}^{1:N},\mathbf{e}_{1:T}^{1:N^2}) \\
    &= \nabla\,{\rm log}\!\left[p(\mathbf{y}_{1}^{1:N}|\mathbf{x}_{1}^{1:N})p(\mathbf{x}_{1}^{1:N}|\mathbf{z}_{1}^{1:N})p(\mathbf{z}_{1}^{1:N})\right] + \sum_{t=2}^{T}\nabla\,{\rm log}\!\left[p(\mathbf{y}_{t}^{1:N}|\mathbf{x}_{t}^{1:N})p(\mathbf{x}_{t}^{1:N}|\mathbf{x}_{t-1}^{1:N},\mathbf{z}_{t}^{1:N})\right]\\ 
    &\,\,\,\,\,\,\,+ \sum_{t=2}^{T}\nabla\,{\rm log}\!\left[p(\mathbf{z}_{t}^{1:N}|\mathbf{z}_{t-1}^{1:N},\mathbf{x}_{t-1}^{1:N},\mathbf{c}_{t-1}^{1:N},\mathbf{e}_{t-1}^{1:N^2})p(\mathbf{e}_{t}^{1:N^2}|\mathbf{e}_{t-1}^{1:N^2},\mathbf{z}_{t}^{1:N},\mathbf{x}_{t}^{1:N})p(\mathbf{c}_{t}^{1:N}|\mathbf{c}_{t-1}^{1:N},\mathbf{z}_{t-1}^{1:N})\right] \\
    &= \nabla\,{\rm log}\!\left[\prod_{n=1}^{N}p(\mathbf{y}_{1}^{n}|\mathbf{x}_{1}^{n})\cdot\prod_{n=1}^{N}p(\mathbf{x}_{1}^{n}|z_{1}^{n})\cdot p(\mathbf{z}_{1}^{1:N})\right] + \sum_{t=2}^{T}\nabla\,{\rm log}\!\left[\prod_{n=1}^{N}p(\mathbf{y}_{t}^{n}|\mathbf{x}_{t}^{n})\cdot \prod_{n=1}^{N}p(\mathbf{x}_{t}^{n}|\mathbf{x}_{t-1}^{n},z_{t}^{n}) \right]\\
    &\,\,\,\,\,\,\,+ \sum_{t=2}^{T}\nabla\,{\rm log}\!\left[ \prod_{n=1}^{N}\!\prod_{m=1}^{N}\!p(z_{t}^{n}|z_{t-1}^{m}, \mathbf{x}_{t-1}^{m,n}, c_{t}^{n}, e_t^{m\rightarrow n}) \cdot \prod_{n=1}^{N}\!\prod_{m=1}^{N}\!p(e_{t}^{m\rightarrow n}|e_{t-1}^{m\rightarrow n}, \mathbf{z}_{t}^{m,n}, \mathbf{x}_{t}^{m,n}) \cdot \prod_{n=1}^{N}p(c_{t}^{n}|c_{t-1}^{n},z_{t-1}^{n})\right]
\end{align*}
where we model the interactions among objects via $p(z_{t}^{n}|z_{t-1}^{m}, \mathbf{x}_{t-1}^{m,n}, c_{t}^{n}, e_t^{m\rightarrow n})$ without instantaneous dependences. Thus, $\nabla{\rm log}\,p(\mathbf{x},\mathbf{y})$ can be written as:
\begin{align*}
    \nabla{\rm log}\,p(\mathbf{x},\mathbf{y})&=\mathbb{E}_{p(\mathbf{z},\mathbf{c},\mathbf{e}|\mathbf{x},\mathbf{y})}\left[\nabla{\rm log}\,p(\mathbf{x},\mathbf{y},\mathbf{z},\mathbf{c},\mathbf{e})\right]\\
    &= \mathbb{E}_{p(\mathbf{z}_{1:T}^{1:N},\mathbf{c}_{1:T}^{1:N},\mathbf{e}_{1:T}^{1:N^2}|\mathbf{x}_{1:T}^{1:N},\mathbf{y}_{1:T}^{1:N})}\left[\nabla{\rm log}\,p(\mathbf{x}_{1:T}^{1:N},\mathbf{y}_{1:T}^{1:N},\mathbf{z}_{1:T}^{1:N},\mathbf{c}_{1:T}^{1:N},\mathbf{e}_{1:T}^{1:N^2})\right] \\
    &= \sum_{\mathbf{k}}p(\mathbf{z}_1^{1:N}=\mathbf{k}|\mathbf{x}_{1:T}^{1:N},\mathbf{y}_{1:T}^{1:N}) \nabla\,{\rm log}\left[\prod_{n=1}^{N}p(\mathbf{y}_{1}^{n}|\mathbf{x}_{1}^{n})\cdot \prod_{n=1}^{N}p(\mathbf{x}_{1}^{n}|\mathbf{z}_{1}^{n})\cdot p(\mathbf{z}_{1}^{1:N}=\mathbf{k})\right] \\
    &\,\,\,\,\,\,\,+ \sum_{t=2}^{T}\sum_{\mathbf{k},\mathbf{j},\mathbf{q},\mathbf{p},\mathbf{s},\mathbf{t}}\!\! \xi(\mathbf{k},\mathbf{j},\mathbf{q},\mathbf{p},\mathbf{s},\!\mathbf{t})\,\nabla\,{\rm log}\!\left[\prod_{n=1}^{N}p(\mathbf{y}_{t}^{n}|\mathbf{x}_{t}^{n})\cdot \prod_{n=1}^{N}p(\mathbf{x}_{t}^{n}|\mathbf{x}_{t-1}^{n},z_{t}^{n}=k^n)\right]\\
    &\,\,\,\,\,\,\,+ \sum_{t=2}^{T}\sum_{\mathbf{k},\mathbf{j},\mathbf{q},\mathbf{p},\mathbf{s},\mathbf{t}}\!\!\xi(\mathbf{k},\mathbf{j},\mathbf{q},\mathbf{p},\mathbf{s},\mathbf{t})\,\nabla\,{\rm log}\left[ \prod_{n=1}^{N}\!\prod_{m=1}^{N}\!\,p(z_{t}^{n}\!=\!k^n|z_{t-1}^{m}\!=\!j^m, \mathbf{x}_{t-1}^{m,n}, c_{t}^{n}\!=\!q^n, e_t^{m\rightarrow n}\!=\!s^{m\rightarrow n})\right]\\
    &\,\,\,\,\,\,\,+ \sum_{t=2}^{T}\sum_{\mathbf{k},\mathbf{j},\mathbf{q},\mathbf{p},\mathbf{s},\mathbf{t}}\!\!\xi(\mathbf{k},\mathbf{j},\mathbf{q},\mathbf{p},\mathbf{s},\mathbf{t})\,\nabla\,{\rm log}\!\left[\prod_{n=1}^{N}\!\prod_{m=1}^{N}\!p(e_{t}^{m\rightarrow n}=s^{m\rightarrow n}|e_{t-1}^{m\rightarrow n}=t^{m\rightarrow n}, \mathbf{z}_{t}^{m,n}=\mathbf{j}^{m,n}, \mathbf{x}_{t}^{m,n})\right]\\
    &\,\,\,\,\,\,\,+ \sum_{t=2}^{T}\sum_{\mathbf{k},\mathbf{j},\mathbf{q},\mathbf{p},\mathbf{s},\mathbf{t}}\!\!\xi(\mathbf{k},\mathbf{j},\mathbf{q},\mathbf{p},\mathbf{s},\mathbf{t})\,\nabla\,{\rm log}\!\left[\prod_{n=1}^{N}p(c_{t}^{n}\!=\!q^n|c_{t-1}^{n}\!=\!p^n,z_{t-1}^{n}\!=\!j^n)\right]\\
    &= \sum_{\mathbf{k}}\gamma(\mathbf{k})\, \nabla\,{\rm log}\!\left[B_1(k^n)\cdot \pi(\mathbf{k})\right] \\
    &\,\,\,\,\,\,\,+\sum_{t=2}^{T}\sum_{\mathbf{k},\mathbf{j},\mathbf{q},\mathbf{p},\mathbf{s},\mathbf{t}}\!\!\xi(\mathbf{k},\mathbf{j},\mathbf{q},\mathbf{p},\mathbf{s},\mathbf{t})\,\nabla\,{\rm log}\!\left[B_t(\mathbf{k})\cdot 
    A_t(\mathbf{k},\mathbf{j},\mathbf{q},\mathbf{s}) \cdot E_t(\mathbf{j},\mathbf{s},\mathbf{t}) \cdot C_t(\mathbf{q},\mathbf{p},\mathbf{j})\right]
\end{align*}
where 
\begin{align*}
\pi(\mathbf{k}) &= p(\mathbf{z}_{1}^{1:N}=\mathbf{k}),\\
\gamma(\mathbf{k}) &= p(\mathbf{z}_1^{1:N}=\mathbf{k}|\mathbf{x}_{1:T}^{1:N},\mathbf{y}_{1:T}^{1:N}),\\
\xi(\mathbf{k},\mathbf{j},\mathbf{q},\mathbf{p},\mathbf{s},\mathbf{t}) &= p(\mathbf{z}_t^{1:N}\!\!=\!\mathbf{k},\mathbf{z}_{t-1}^{1:N}\!\!=\!\mathbf{j},\mathbf{c}_t^{1:N}\!=\!\mathbf{q},\mathbf{c}_{t-1}^{1:N}\!=\!\mathbf{p},\mathbf{e}_t^{1:N^2}\!\!=\!\mathbf{s},\mathbf{e}_{t-1}^{1:N^2}\!\!=\!\mathbf{t}|\mathbf{x}_{1:T}^{1:N},\mathbf{y}_{1:T}^{1:N}), \\
B_t(\mathbf{k}) &= \prod_{n=1}^{N}p(\mathbf{y}_{t}^{n}|\mathbf{x}_{t}^{n})\cdot \prod_{n=1}^{N}p(\mathbf{x}_{t}^{n}|\mathbf{x}_{t-1}^{n},z_{t}^{n}=k^n),\\
A_t(\mathbf{k},\mathbf{j},\mathbf{q},\mathbf{s}) &= \prod_{n=1}^{N}\!\prod_{m=1}^{N}\!\,p(z_{t}^{n}\!=\!k^n|z_{t-1}^{m}\!=\!j^m, \mathbf{x}_{t-1}^{m,n}, c_{t}^{n}\!=\!q^n, e_t^{m\rightarrow n}\!=\!s^{m\rightarrow n}),\\
E_t(\mathbf{j},\mathbf{s},\mathbf{t}) &= \prod_{n=1}^{N}\!\prod_{m=1}^{N}\!p(e_{t}^{m\rightarrow n}=s^{m\rightarrow n}|e_{t-1}^{m\rightarrow n}=t^{m\rightarrow n}, \mathbf{z}_{t}^{m,n}=\mathbf{j}^{m,n}, \mathbf{x}_{t}^{m,n})\\
C_t(\mathbf{q},\mathbf{p},\mathbf{j}) &= \prod_{n=1}^{N}p(c_{t}^{n}\!=\!q^n|c_{t-1}^{n}\!=\!p^n,z_{t-1}^{n}\!=\!j^n).
\end{align*}
$\pi(\mathbf{k})$ is the initial joint discrete mode probability. $\prod_{n=1}^{N}p(\mathbf{y}_{1}^{n}|\mathbf{x}_{1}^{n})$ and $\prod_{n=1}^{N}p(\mathbf{y}_{t}^{n}|\mathbf{x}_{t}^{n})$ are emission probability. $\prod_{n=1}^{N}p(\mathbf{x}_{1}^{n}|z_{1}^{n})$ and $\prod_{n=1}^{N}p(\mathbf{x}_{t}^{n}|\mathbf{x}_{t-1}^{n},z_{t}^{n}\!=\!k^n)$  are continuous state transition probability conditioned on different types of discrete modes $k^n$. $A_t(\mathbf{k},\mathbf{j},\mathbf{q},\mathbf{s})$ is the discrete mode transition probability. Besides, $p(\mathbf{z}_1^{1:N}=\mathbf{k}|\mathbf{x}_{1:T}^{1:N},\mathbf{y}_{1:T}^{1:N})$ and $\xi(\mathbf{k},\mathbf{j},\mathbf{q},\mathbf{p},\mathbf{s})$ can be calculated similarly to the forward and backward algorithm in HMMs \citep{collins2013forward}, which is detailed in the next section.

\subsubsection{Forward and Backward Algorithm}\label{app:fwbw}
In this section, we aim at calculating the posterior probability of discrete mode, count, and edge variables $\mathbf{z}$, $\mathbf{c}$, and $\mathbf{e}$ conditioned on observation $\mathbf{y}$ and approximate continuous state $\mathbf{x}$:
\begin{align*}
    p(\mathbf{z}_t,\mathbf{c}_t,\mathbf{e}_t|\mathbf{x}_{1:T},\mathbf{y}_{1:T}) & \propto p(\mathbf{z}_t,\mathbf{c}_t,\mathbf{e}_t,\mathbf{x}_{1:T},\mathbf{y}_{1:T}) \\
    &= \underbrace{p(\mathbf{z}_t,\mathbf{c}_t,\mathbf{e}_t,\mathbf{x}_{1:t},\mathbf{y}_{1:t})}_{Forward}\underbrace{p(\mathbf{x}_{t+1:T},\mathbf{y}_{t+1:T}|\mathbf{x}_{t},\mathbf{z}_t,\mathbf{c}_t,\mathbf{e}_t)}_{Backward} \\
    &= \alpha_t(\mathbf{z}_t,\mathbf{c}_t)\cdot\beta_t(\mathbf{z}_t,\mathbf{c}_t).
\end{align*}
The forward part $\alpha_t(\mathbf{z}_t,\mathbf{c}_t)$ can be expanded as:
\begin{align*}
    \alpha_1(\mathbf{z}_1, \mathbf{c}_1) 
    &= p(\mathbf{z}_1, \mathbf{c}_1,\mathbf{e}_1, \mathbf{x}_1, \mathbf{y}_1) \\
    &= 
    p(\mathbf{z}_1^{1:N}, \mathbf{c}_1^{1:N},\mathbf{e}_1^{1:N^2}, \mathbf{x}_1^{1:N}, \mathbf{y}_1^{1:N})\\
    &= \delta_{\mathbf{c}_1^{1:N}=1} p(\mathbf{z}_1^{1:N})p(\mathbf{e}_1^{1:N^2})p(\mathbf{x}_{1}^{1:N}|\mathbf{z}_1^{1:N})p(\mathbf{y}_{1}^{1:N}|\mathbf{x}_{1}^{1:N}) \\
    &=  \delta_{\mathbf{c}_1^{1:N}=1} p(\mathbf{z}_1^{1:N}) p(\mathbf{e}_1^{1:N^2})\prod_{n=1}^{N}p(\mathbf{x}_{1}^{n}|\mathbf{z}_{1}^{n})\prod_{n=1}^{N}p(\mathbf{y}_{1}^{n}|\mathbf{x}_{1}^{n})\\ 
    \underline{\alpha_t(\mathbf{z}_t, \mathbf{c}_t)} &= p(\mathbf{z}_t,\mathbf{c}_t,\mathbf{e}_t,\mathbf{x}_{1:t},\mathbf{y}_{1:t}) \\
    &= p(\mathbf{z}_t^{1:N},\mathbf{c}_t^{1:N},\mathbf{e}_t^{1:N^2},\mathbf{x}_{1:t}^{1:N},\mathbf{y}_{1:t}^{1:N})\\
    &= \sum_{\mathbf{z}_{t-1}^{1:N},\mathbf{c}_{t-1}^{1:N}}p(\mathbf{z}_t^{1:N},\mathbf{c}_t^{1:N},\mathbf{e}_t^{1:N^2},\mathbf{x}_{1:t}^{1:N},\mathbf{y}_{1:t}^{1:N}, \mathbf{z}_{t-1}^{1:N},\mathbf{c}_{t-1}^{1:N}) \\
    &= \sum_{\mathbf{z}_{t-1}^{1:N},\mathbf{c}_{t-1}^{1:N}}p(\mathbf{z}_{t-1}^{1:N},\mathbf{c}_{t-1}^{1:N},\mathbf{e}_{t-1}^{1:N^2},\mathbf{x}_{1:t-1}^{1:N},\mathbf{y}_{1:t-1}^{1:N})p(\mathbf{c}_{t}^{1:N}|\mathbf{c}_{t-1}^{1:N},\mathbf{z}_{t-1}^{1:N})p(\mathbf{z}_{t}^{1:N}|\mathbf{z}_{t-1}^{1:N},\mathbf{x}_{t-1}^{1:N},\mathbf{c}_{t-1}^{1:N},\mathbf{e}_{t-1}^{1:N^2}) \\
    & \quad\quad\quad\quad\quad\,\, \cdot p(\mathbf{e}_{t}^{1:N^2}|\mathbf{e}_{t-1}^{1:N^2},\mathbf{z}_{t}^{1:N},\mathbf{x}_{t}^{1:N})p(\mathbf{x}_{t}^{1:N}|\mathbf{x}_{t-1}^{1:N},\mathbf{z}_{t}^{1:N})p(\mathbf{y}_{t}^{1:N}|\mathbf{x}_{t}^{1:N}) \\
    & = \sum_{\mathbf{z}_{t-1}^{1:N},\mathbf{c}_{t-1}^{1:N}}\underline{\alpha_{t-1}(\mathbf{z}_{t-1},\mathbf{c}_{t-1})}\prod_{n=1}^{N}p(c_{t}^{n}|c_{t-1}^{n},z_{t-1}^{n})\prod_{n=1}^{N}\!\prod_{m=1}^{N}\!p(z_{t}^{n}|z_{t-1}^{m}, \mathbf{x}_{t-1}^{m,n}, c_{t}^{n}, e_t^{m\rightarrow n}) \\
    & \quad\quad\quad\quad\quad\,\, \cdot \prod_{n=1}^{N}\!\prod_{m=1}^{N}\!p(e_{t}^{m\rightarrow n}|e_{t-1}^{m\rightarrow n}, \mathbf{z}_{t}^{m,n}, \mathbf{x}_{t}^{m,n}) \prod_{n=1}^{N}p(\mathbf{x}_{t}^{n}|\mathbf{x}_{t-1}^{n},z_{t}^{n}) \prod_{n=1}^{N}p(\mathbf{y}_{t}^{n}|\mathbf{x}_{t}^{n}), 
\end{align*}
where $\alpha_t(\mathbf{z}_t,\mathbf{c}_t)$ can be expressed by $\alpha_{t-1}(\mathbf{z}_{t-1},\mathbf{c}_{t-1})$ recursively with variable transitions and emissions.

The backward part $\beta_t(\mathbf{z}_t,\mathbf{c}_{t})$ can be expanded as:
\begin{align*}
    \beta_T(\mathbf{z}_T,\mathbf{c}_T) &= 1 \\
    \underline{\beta_t(\mathbf{z}_t,\mathbf{c}_t)} &= p(\mathbf{x}_{t+1:T},\mathbf{y}_{t+1:T}|\mathbf{x}_{t},\mathbf{z}_t,\mathbf{c}_t,\mathbf{e}_t) \\
    &= p(\mathbf{x}_{t+1:T}^{1:N},\mathbf{y}_{t+1:T}^{1:N}|\mathbf{x}_{t}^{1:N}, \mathbf{z}_{t}^{1:N}, \mathbf{c}_{t}^{1:N}, \mathbf{e}_{t}^{1:N^2})\\
    &= \sum_{\mathbf{z}_{t+1}^{1:N},\mathbf{c}_{t+1}^{1:N}}p(\mathbf{x}_{t+1:T}^{1:N},\mathbf{y}_{t+1:T}^{1:N},\mathbf{z}_{t+1}^{1:N},\mathbf{c}_{t+1}^{1:N} |\mathbf{x}_{t}^{1:N}, \mathbf{z}_{t}^{1:N}, \mathbf{c}_{t}^{1:N}, \mathbf{e}_{t}^{1:N^2})\\
    &= \sum_{\mathbf{z}_{t+1}^{1:N},\mathbf{c}_{t+1}^{1:N}} p(\mathbf{c}_{t+1}^{1:N}|\mathbf{c}_{t}^{1:N},\mathbf{z}_{t}^{1:N})p(\mathbf{z}_{t+1}^{1:N}|\mathbf{z}_{t}^{1:N},\mathbf{x}_{t}^{1:N},\mathbf{c}_{t}^{1:N},\mathbf{e}_{t}^{1:N^2}) \\
    & \cdot p(\mathbf{x}_{t+1}^{1:N}|\mathbf{x}_{t}^{1:N},\mathbf{z}_{t+1}^{1:N})p(\mathbf{e}_{t+1}^{1:N^2}|\mathbf{e}_{t}^{1:N^2},\mathbf{z}_{t+1}^{1:N},\mathbf{x}_{t+1}^{1:N})p(\mathbf{y}_{t+1}^{1:N}|\mathbf{x}_{t+1}^{1:N})p(\mathbf{x}_{t+2:T}^{1:N},\mathbf{y}_{t+2:T}^{1:N}|\mathbf{x}_{t+1}^{1:N}, \mathbf{z}_{t+1}^{1:N}, \mathbf{c}_{t+1}^{1:N}, \mathbf{e}_{t+1}^{1:N^2}) \\
    &= \sum_{\mathbf{z}_{t+1}^{1:N},\mathbf{c}_{t+1}^{1:N}} \prod_{n=1}^{N}p(c_{t+1}^{n}|c_{t}^{n},z_{t}^{n})
    \prod_{n=1}^{N}\!\prod_{m=1}^{N}\!p(z_{t+1}^{n}|z_{t}^{m}, \mathbf{x}_{t}^{m,n}, c_{t+1}^{n}, e_{t+1}^{m\rightarrow n}) \\
    & \cdot \prod_{n=1}^{N}p(\mathbf{x}_{t+1}^{n}|\mathbf{x}_{t}^{n},z_{t+1}^{n})
    \prod_{n=1}^{N}\!\prod_{m=1}^{N}\!p(e_{t+1}^{m\rightarrow n}|e_{t}^{m\rightarrow n}, \mathbf{z}_{t+1}^{m,n}, \mathbf{x}_{t+1}^{m,n})  \prod_{n=1}^{N}p(\mathbf{y}_{t+1}^{n}|\mathbf{x}_{t+1}^{n})\,\,\underline{\beta_{t+1}(\mathbf{z}_{t+1},\mathbf{c}_{t+1})},
\end{align*}
where $\beta_t(\mathbf{z}_t,\mathbf{c}_t)$ can be computed via $\beta_{t+1}(\mathbf{z}_{t+1},\mathbf{c}_{t+1})$ recursively with variable transitions and emissions.

\subsection{Further Model Interactions between Continuous Variables}
In the main paper, we model interactions between objects by dependence on discrete mode variables only. This means that based on the derived discrete mode transition, the continuous state transition $p(\mathbf{x}_t^n|\mathbf{x}_{t-1}^n, z_t^n)$ and observation emission $p(\mathbf{y}_t^n|\mathbf{x}_t^n)$ are per-object dynamics only without interactions. However, in some real-world scenarios, the interactions between objects also happen to continuous variables. For example, in each motion type, object A still influences the detailed motion of object B. We show some preliminary results in this section and leave more comprehensive experiments as future work.

\section{More Experiments}

\subsection{New splitting and larger ODE-driven particle datasets}
\label{appendix: New split and larger ODE}
In our original ODE-driven particle dataset we used around 5k samples for training, around 200 samples for validation and testing. We tested the scalability of our method in terms of scaling to one larger (approximately 20x larger) dataset. The original dataset takes ~37,000 epoches to achieve convergence and the final performance of our GRASS model is: 0.528, 0.519, 0.794, and 0.790 for NMI, ARI, Accuracy, and F1, respectively.
The 20x larger dataset takes ~39,000 epochs and the final performance of our GRASS model is 0.525, 0.531, 0.814, and 0.802. We find the training time before convergence and the performance of our model are almost the same, which shows the scalability of our method to larger datasets.

The splitting strategy of the synthesized dataset follows the recent SOTA method, REDSDS~\citep{ansari2021deep}. REDSDS has 10,000 and 500 samples for training and testing of the 3-mode system (test data is around 5\% of training data). We follow the proportion and have 4,928 samples for training and 204 samples for testing (around 5\%). For the ODE-driven particle dataset, we also conduct a new splitting (4200/420/420 for training/validation/testing). The results of our GRASS model on the new splitting dataset are 0.522, 0.518, 0.809, and 0.805 for NMI, ARI, Accuracy, and F1, respectively, which shows almost the same performance as the original splitting in the main paper.

\subsection{Ablation studies with standard derivations}
\label{appendix: Ablation studies with error bars}
Ablations studies with standard derivations are in 
Tables~\ref{appendix: Number of Particles table},~\ref{appendix: Number of interactions table},~\ref{appendix: no interaction table},~and \ref{appendix: Number of Modes table}. We can see that the conclusions remain the same as in the main paper for ablation studies of different numbers of objects, different numbers of interactions, with or without interactions, and different numbers of predefined modes. Note that in Table~\ref{appendix: Number of Particles table} and Table~\ref{appendix: Number of interactions table}, we can see that with different number of objects or interactions, GRASS has consistently better performance with the lowest variances. 
\begin{table*}[t]
	\centering
	\scriptsize
	\tabcolsep=0.100cm
	\setlength\arrayrulewidth{1.0pt}
	\caption{Analyses on different numbers of objects on ODE-driven Particle dataset, while \emph{increasing} the average number of interactions per object per time series, i.e, 2.3 interactions for 3 particles, 6.1 for 5, and 12.5 for 10. */* denotes NMI / $F_1$.}
	\vspace{4pt}
	\begin{tabular}{lccc}
		\toprule
  Number of Particles & 3 & 5 & 10 \\
		\midrule
        rSLDS & 0.257$\pm$0.023 / 0.443$\pm$0.041 & 0.252$\pm$0.033 / 0.437$\pm$0.039 & 0.246$\pm$0.027 / 0.430$\pm$0.045 \cr
        SNLDS & 0.368$\pm$0.027 / 0.664$\pm$0.053 & 0.361$\pm$0.031 / 0.656$\pm$0.042 & 0.354$\pm$0.035 / 0.651$\pm$0.059 \cr
        REDSDS & 0.418$\pm$0.016 / 0.701$\pm$0.027 & 0.411$\pm$0.023 / 0.692$\pm$0.029 & 0.405$\pm$0.024 / 0.687$\pm$0.022 \cr
        MOSDS (this paper) & 0.469$\pm$0.020 / 0.757$\pm$0.032  & 0.461$\pm$0.024 / 0.752$\pm$0.027  & 0.456$\pm$0.029 /  
 0.748$\pm$0.035 \cr
        GRASS (this paper) & \textbf{0.528$\pm$0.014} / \textbf{0.790$\pm$0.021} & \textbf{0.524$\pm$0.019} / \textbf{0.784$\pm$0.025} & \textbf{0.519$\pm$0.021} / \textbf{0.781$\pm$0.018} \cr
		\bottomrule
	\end{tabular}
	\label{appendix: Number of Particles table}
\end{table*}

\begin{table*}[t]
	\centering
	\scriptsize
	\tabcolsep=0.100cm
	\setlength\arrayrulewidth{1.0pt}
	\caption{Analyses on different numbers of objects on ODE-driven Particle, while \emph{fixing} the average number of interactions per object per time series, i.e, 2.3 interactions. */* denotes NMI / $F_1$.}
	\vspace{4pt}
	\begin{tabular}{lccc}
		\toprule
  Number of Particles & 3 & 5 & 10 \\
		\midrule
        rSLDS & 0.257$\pm$0.023 / 0.443$\pm$0.041 & 0.262$\pm$0.034 / 0.444$\pm$0.037 & 0.253$\pm$0.028 / 0.437$\pm$0.042 \cr
        SNLDS & 0.368$\pm$0.027 / 0.664$\pm$0.053 & 0.365$\pm$0.030 / 0.666$\pm$0.047 & 0.362$\pm$0.028 / 0.659$\pm$0.051 \cr
        REDSDS & 0.418$\pm$0.016 / 0.701$\pm$0.027 & 0.423$\pm$0.023 / 0.706$\pm$0.031 & 0.413$\pm$0.022 / 0.694$\pm$0.028\cr
        MOSDS (this paper) & 0.469$\pm$0.020 / 0.757$\pm$0.032 & 0.471$\pm$0.025 / 0.763$\pm$0.036 & 0.464$\pm$0.021 / 0.754$\pm$0.035\cr
        GRASS (this paper) & \textbf{0.528$\pm$0.014} / \textbf{0.790$\pm$0.021} & \textbf{0.530$\pm$0.012} / \textbf{0.792$\pm$0.019} & \textbf{0.524$\pm$0.017} / \textbf{0.786$\pm$0.024}\cr
		\bottomrule
	\end{tabular}
	\label{appendix: Number of interactions table}
\end{table*}

\begin{table*}[t]
	\centering
	\scriptsize
	\tabcolsep=0.100cm
	\setlength\arrayrulewidth{1.0pt}
	\caption{Analyses of robustness to datasets without interactions on ODE-driven Particle dataset. */* denotes NMI / $F_1$.}
	\vspace{4pt}
	\begin{tabular}{lcc}
		\toprule
  Method & dataset w/ interaction & dataset w/o interaction  \\
		\midrule
        rSLDS & 0.257$\pm$0.023 / 0.443$\pm$0.041 &  0.471$\pm$0.024 / 0.686$\pm$0.035 \cr
        SNLDS & 0.368$\pm$0.027 / 0.664$\pm$0.053 & 0.534$\pm$0.032 / 0.772$\pm$0.046 \cr
        REDSDS & 0.418$\pm$0.016 / 0.701$\pm$0.027 & \textbf{0.579$\pm$0.013} / \textbf{0.838$\pm$0.022} \cr
        MOSDS (this paper) & 0.469$\pm$0.020 / 0.757$\pm$0.032  & 0.563$\pm$0.027 / 0.817$\pm$0.039 \cr
        GRASS (this paper) & \textbf{0.528$\pm$0.014} / \textbf{0.790$\pm$0.021} & 0.573$\pm$0.008 / 0.826$\pm$0.018 \cr
		\bottomrule
	\end{tabular}
	\label{appendix: no interaction table}
\end{table*}

\begin{table*}[t]
	\centering
	\scriptsize
	\tabcolsep=0.100cm
	\setlength\arrayrulewidth{1.0pt}
	\caption{Analyses on robustness to different maximal numbers of predefined modes. */* denotes NMI / $F_1$.}
	\vspace{4pt}
	\begin{tabular}{lccc}
		\toprule
  Number of Modes & 3 & 5 & 10 \\
		\midrule
        rSLDS & 0.257$\pm$0.023 / 0.443$\pm$0.041 & 0.253$\pm$0.025 / 0.438$\pm$0.043 & 0.248$\pm$0.032 / 0.436$\pm$0.047\cr
        SNLDS & 0.368$\pm$0.027 / 0.664$\pm$0.053 & 0.365$\pm$0.032 / 0.661$\pm$0.047 & 0.362$\pm$0.036 / 0.657$\pm$0.059 \cr
        REDSDS & 0.418$\pm$0.016 / 0.701$\pm$0.027 & 0.415$\pm$0.023 / 0.696$\pm$0.035 & 0.413$\pm$0.026 / 0.694$\pm$0.031 \cr
        MOSDS (this paper) & 0.469$\pm$0.020 / 0.757$\pm$0.032  & 0.466$\pm$0.028 / 0.759$\pm$0.037  & 0.462$\pm$0.033 / 0.754$\pm$0.042\cr
        GRASS (this paper) &\textbf{0.528$\pm$0.014} / \textbf{0.790$\pm$0.021} & \textbf{0.532$\pm$0.020} / \textbf{0.794$\pm$0.025} & \textbf{0.527$\pm$0.022} / \textbf{0.784$\pm$0.026} \cr 
		\bottomrule
	\end{tabular}
	\label{appendix: Number of Modes table}
\end{table*}

\end{document}